\documentclass[10pt,twocolumn,letterpaper]{article}

\usepackage[pagenumbers]{cvpr} 

\usepackage{graphicx}
\usepackage{amsmath}
\usepackage{amssymb}
\usepackage{booktabs}
\usepackage{lipsum}
\usepackage[dvipsnames]{xcolor}
\usepackage{physics}
\usepackage{dsfont}
\usepackage{multirow}
\usepackage{balance}

\usepackage[pagebackref,breaklinks,colorlinks]{hyperref}

\PassOptionsToPackage{sort,comma,numbers}{natbib}
\usepackage[capitalize]{cleveref}
\crefname{section}{Sec.}{Secs.}
\Crefname{section}{Section}{Sections}
\Crefname{table}{Table}{Tables}
\crefname{table}{Tab.}{Tabs.}

\def\cvprPaperID{3500} 
\def\confName{CVPR}
\def\confYear{2023}
\begin{document}

\title{Scaling up GANs for Text-to-Image Synthesis}

\author{Minguk Kang$^{1,3}$
\hspace{8mm}
Jun-Yan Zhu$^{2}$
\hspace{8mm}
Richard Zhang$^{3}$ \\
\hspace{8mm}
Jaesik Park$^{1}$ \vspace{1mm}
\hspace{8mm}
Eli Shechtman$^{3}$
\hspace{8mm}
Sylvain Paris$^{3}$ 
\hspace{8mm}
Taesung Park$^{3}$
\vspace{2.5mm}\\
$^{1}$POSTECH
\hspace{12mm}
$^{2}$Carnegie Mellon University
\hspace{12mm}
$^{3}$Adobe Research
}

\maketitle
\newcommand{\minguk}[1]{{\textcolor{red}{MinGuk: #1}}}
\newcommand{\jaesik}[1]{{\textcolor{violet}{Jaesik: #1}}}
\newcommand{\tae}[1]{{\textcolor{red}{Taesung: #1}}}

\definecolor{darkblue}{HTML}{202282} 
\definecolor{lightgray}{HTML}{E1E1E1} 
\newcommand{\newtext}[1]{{\textcolor{violet}{#1}}}

\newcommand{\perm}{\boldsymbol{\pi}}

\newcommand{\myparagraph}[1]{\vspace{-5pt}\paragraph{#1}}
\newcommand{\mysubsection}[1]{\vspace{1mm}\noindent{\bf #1}}

\newcommand{\cmark}{\ding{51}}
\newcommand{\xmark}{\ding{55}}

\def\L{\mathcal{L}}

\def\vpos{\bm{v^{+}}}
\def\vneg{\bm{v^{-}}}

\def\X{\mathcal{X}}
\def\Y{\mathcal{Y}}
\def\Z{\mathcal{Z}}
\def\data{\mathcal{D}}
\def\P{\mathbb{P}}
\def\Px{\mathbb{P}_{\mathcal{X}}}
\def\Pg{\mathbb{P}_{\theta}}
\def \Lv{\mathcal{L}_{\text{models}}}
\def \kmax{K\max}

\def\xhat{{\hat{\bm x}}}
\def\yhat{{\hat{\bm y}}}
\def\zhat{{\hat{\bm z}}}

\def\xtilde{{\tilde{\x}}}
\def\ytilde{{\tilde{\y}}}
\newcommand*\wc{{\mkern 2mu\cdot\mkern 2mu}}

\def\ztilde{{\tilde{\z}}}

\def\xbar{{\bar{\x}}}
\def\ybar{{\bar{\y}}}
\def\zbar{{\bar{\z}}}

\def\xstar{{\x^{*}}}
\def\ystar{{\y^{*}}}
\def\zstar{{\z^{*}}}

\def\D{{D}}
\def\G{{G}}
\def\Re{\mathds{R}}
\newcommand{\expect}[1]{\mathbb{E}_{#1}}
\def\NCE{\ell}

\newcommand{\citeColored}[2]{{\hypersetup{citecolor=#1}\cite{#2}}}

\newcommand{\fid}{Fr\'echet Inception Distance\xspace}

\def\rvepsilon{{\mathbf{\epsilon}}}
\def\rvtheta{{\mathbf{\theta}}}
\def\rva{{\mathbf{a}}}
\def\rvb{{\mathbf{b}}}
\def\rvc{{\mathbf{c}}}
\def\rvd{{\mathbf{d}}}
\def\rve{{\mathbf{e}}}
\def\rvf{{\mathbf{f}}}
\def\rvg{{\mathbf{g}}}
\def\rvh{{\mathbf{h}}}
\def\rvu{{\mathbf{i}}}
\def\rvj{{\mathbf{j}}}
\def\rvk{{\mathbf{k}}}
\def\rvl{{\mathbf{l}}}
\def\rvm{{\mathbf{m}}}
\def\rvn{{\mathbf{n}}}
\def\rvo{{\mathbf{o}}}
\def\rvp{{\mathbf{p}}}
\def\rvq{{\mathbf{q}}}
\def\rvr{{\mathbf{r}}}
\def\rvs{{\mathbf{s}}}
\def\rvt{{\mathbf{t}}}
\def\rvu{{\mathbf{u}}}
\def\rvv{{\mathbf{v}}}
\def\rvw{{\mathbf{w}}}
\def\rvx{{\mathbf{x}}}
\def\rvy{{\mathbf{y}}}
\def\rvz{{\mathbf{z}}}

\newcommand{\seg}{\rvs_c}
\newcommand{\repr}{\rvr}
\newcommand{\repru}{\repr_{u, \sP}}
\newcommand{\uprepru}{\repr_{u, \sP}^{\uparrow}}
\newcommand{\f}{f}
\newcommand{\h}{h}

\newcommand{\figleft}{{\em (Left)}}
\newcommand{\figcenter}{{\em (Center)}}
\newcommand{\figright}{{\em (Right)}}
\newcommand{\figtop}{{\em (Top)}}
\newcommand{\figbottom}{{\em (Bottom)}}
\newcommand{\captiona}{{\em (a)}}
\newcommand{\captionb}{{\em (b)}}
\newcommand{\captionc}{{\em (c)}}
\newcommand{\captiond}{{\em (d)}}

\newcommand{\newterm}[1]{{\bf #1}}

\newcommand{\reffig}[1]{Figure~\ref{fig:#1}}
\newcommand{\refsec}[1]{Section~\ref{sec:#1}}
\newcommand{\refapp}[1]{Appendix~\ref{sec:#1}}
\newcommand{\reftbl}[1]{Table~\ref{tbl:#1}}
\newcommand{\refalg}[1]{Algorithm~\ref{alg:#1}}
\newcommand{\refline}[1]{Line~\ref{line:#1}}
\newcommand{\shortrefsec}[1]{\S~\ref{sec:#1}}
\newcommand{\refeq}[1]{Eqn.~\ref{eq:#1}}
\newcommand{\refeqshort}[1]{(\ref{eq:#1})}
\newcommand{\shortrefeq}[1]{\ref{eq:#1}}
\newcommand{\lblfig}[1]{\label{fig:#1}}
\newcommand{\lblsec}[1]{\label{sec:#1}}
\newcommand{\lbleq}[1]{\label{eq:#1}}
\newcommand{\lbltbl}[1]{\label{tbl:#1}}
\newcommand{\lblalg}[1]{\label{alg:#1}}
\newcommand{\lblline}[1]{\label{line:#1}}
\newcommand{\ignorethis}[1]{}
\newcommand{\revision}[1]{\color{black}#1\color{black}}
\newcommand{\myitem}{\vspace{-5pt}\item}

\def\ceil#1{\lceil #1 \rceil}
\def\floor#1{\lfloor #1 \rfloor}
\def\1{\bm{1}}
\newcommand{\train}{\mathcal{D}}
\newcommand{\valid}{\mathcal{D_{\mathrm{valid}}}}
\newcommand{\test}{\mathcal{D_{\mathrm{test}}}}

\def\eps{{\epsilon}}

\newcommand{\image}{{\rvx}}
\newcommand{\latent}{{\rvz}}
\newcommand{\images}{{\mathcal{X}}}
\newcommand{\imagedist}{{p_{\text{data}}(\image)}}
\newcommand{\latentdist}{{p(\latent)}}
\newcommand{\imageD}{{D_X}}
\newcommand{\Fnet}{{F}}
\newcommand{\sketch}{{\rvy}}
\newcommand{\sketches}{{\mathcal{Y}}}
\newcommand{\sketchdist}{{p_{\text{data}}(\sketch)}}
\newcommand{\sketchD}{{D_Y}}
\newcommand{\modelold}{{G(\rvz; \theta)}}
\newcommand{\modelnew}{{G(\rvz; \theta')}}
\newcommand{\losssketch}{{\mathcal{L}_{\text{sketch}}}}
\newcommand{\lossimage}{{\mathcal{L}_{\text{image}}}}
\newcommand{\lossweight}{{\mathcal{L}_{\text{weight}}}}

\newcommand{\method}{{GAN Sketching}}

\newcommand{\pdata}{{D}}
\newcommand{\ptrain}{\hat{p}_{\rm{data}}}
\newcommand{\Ptrain}{\hat{P}_{\rm{data}}}
\newcommand{\pmodel}{p_{\rm{model}}}
\newcommand{\Pmodel}{P_{\rm{model}}}
\newcommand{\ptildemodel}{\tilde{p}_{\rm{model}}}
\newcommand{\pencode}{p_{\rm{encoder}}}
\newcommand{\pdecode}{p_{\rm{decoder}}}
\newcommand{\precons}{p_{\rm{reconstruct}}}

\newcommand{\laplace}{\mathrm{Laplace}} %

\newcommand{\Ls}{\mathcal{L}}
\newcommand{\R}{\mathbb{R}}
\newcommand{\emp}{\tilde{p}}
\newcommand{\lr}{\alpha}
\newcommand{\reg}{\lambda}
\newcommand{\rect}{\mathrm{rectifier}}
\newcommand{\softmax}{\mathrm{softmax}}
\newcommand{\sigmoid}{\sigma}
\newcommand{\softplus}{\zeta}
\newcommand{\Var}{\mathrm{Var}}
\newcommand{\standarderror}{\mathrm{SE}}
\newcommand{\Cov}{\mathrm{Cov}}
\newcommand{\normlzero}{L^0}
\newcommand{\normlone}{L^1}
\newcommand{\normltwo}{L^2}
\newcommand{\normlp}{L^p}
\newcommand{\normmax}{L^\infty}

\newcommand{\parents}{Pa} %

\newcommand{\xpar}[1]{\noindent\textbf{#1}\ \ }
\newcommand{\vpar}[1]{\vspace{3mm}\noindent\textbf{#1}\ \ }

\newcommand{\shapenet}{ShapeNet\xspace}
\newcommand{\pascal}{PASCAL 3D+\xspace}

\newcommand{\degree}{\ensuremath{^\circ}\xspace}
\newcommand{\ignore}[1]{}

\newcommand{\fcseven}{$\mbox{fc}_7$}

\renewcommand*{\thefootnote}{\arabic{footnote}}

\def\naive{na\"{\i}ve\xspace}
\def\Naive{Na\"{\i}ve\xspace}

\makeatletter
\DeclareRobustCommand\onedot{\futurelet\@let@token\@onedot}
\def\@onedot{\ifx\@let@token.\else.\null\fi\xspace}
\def\eg{e.g\onedot,\xspace} 
\def\Eg{E.g\onedot,}
\def\ie{i.e\onedot,\xspace} 
\def\Ie{\emph{I.e}\onedot,}
\def\cf{\emph{c.f}\onedot} \def\Cf{\emph{C.f}\onedot}
\def\etc{\emph{etc}\onedot} \def\vs{\emph{vs}\onedot}
\def\wrt{w.r.t\onedot} \def\dof{d.o.f\onedot}
\def\etal{\emph{et al}\onedot}
\makeatother

\newcommand*{\img}[1]{%
    \raisebox{-.25\baselineskip}{%
        \includegraphics[
        height=\baselineskip,
        width=\baselineskip,
        keepaspectratio,
        ]{#1}%
    }%
}

\fboxsep=0mm%
\fboxrule=2pt%
\newcommand*{\myoverpic}[2]{
  \begin{overpic}[width=.18\linewidth]{#1}
     \put(60,63){\fbox{\includegraphics[width=.06\linewidth]{#2}}}  
  \end{overpic}
}

\newcommand{\OursAcronym}{GigaGAN}

\newcommand{\figwidth}{0.98\linewidth}

\begin{abstract}
The recent success of text-to-image synthesis has taken the world by storm and captured the general public's imagination. From a technical standpoint, it also marked a drastic change in the favored architecture to design generative image models. GANs used to be the de facto choice, with techniques like StyleGAN. With DALL$\cdot$E~2, auto\-regressive and diffusion models became the new standard for large-scale generative models overnight. This rapid shift raises a fundamental question: can we scale up GANs to benefit from large datasets like LAION? We find that na\"ively increasing the capacity of the StyleGAN architecture quickly becomes unstable. We introduce GigaGAN, a new GAN architecture that far exceeds this limit, demonstrating GANs as a viable option for text-to-image synthesis. GigaGAN offers three major advantages. First, it is orders of magnitude faster at inference time, taking only 0.13 seconds to synthesize a 512px image. Second, it can synthesize high-resolution images, for example, 16-megapixel images in 3.66 seconds. Finally, GigaGAN supports various latent space editing applications such as latent interpolation, style mixing, and vector arithmetic operations.
\end{abstract}
\section{Introduction}

\begin{figure*}[t!]
    \centering
    \includegraphics[width=0.98\linewidth]{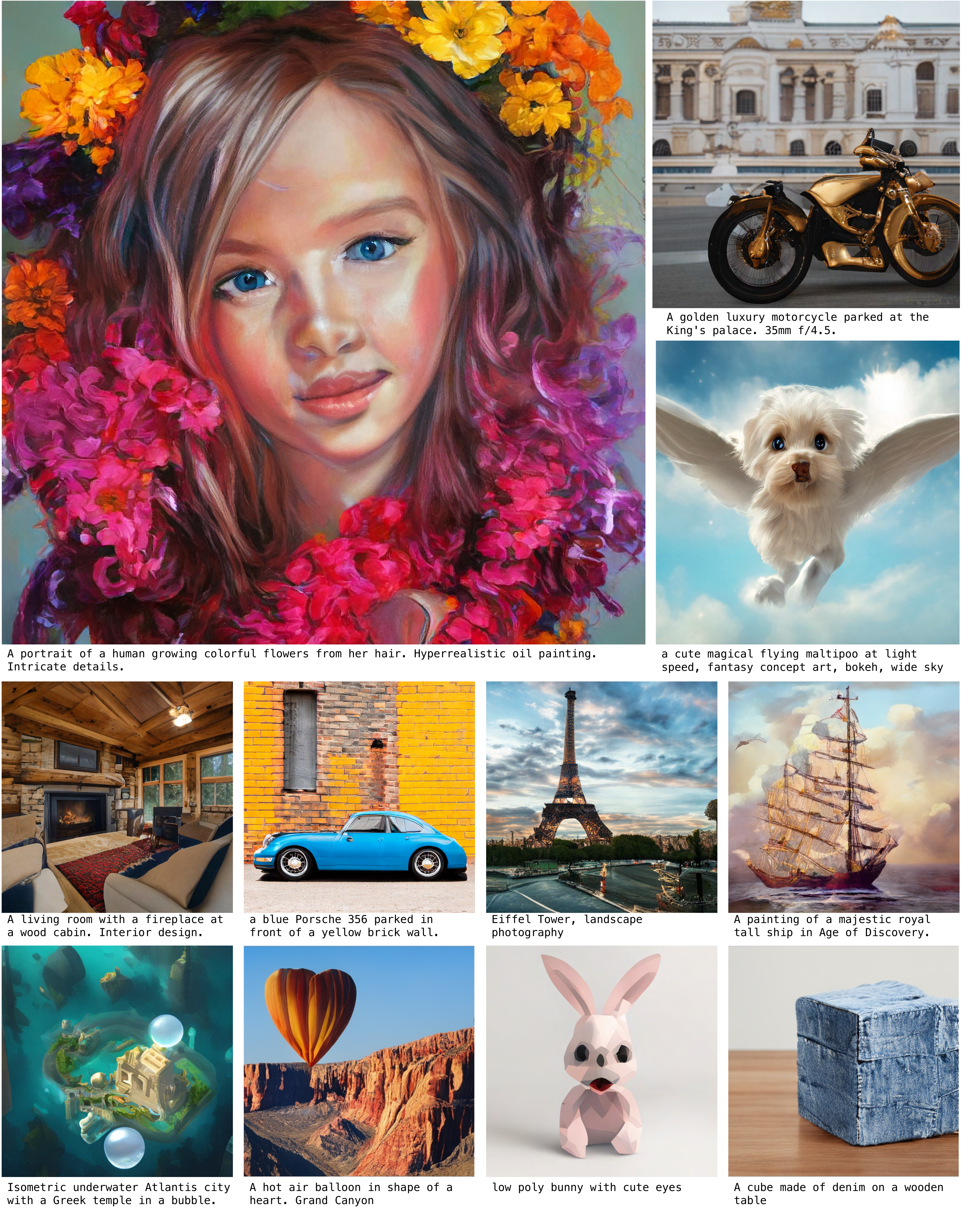}
    \vspace{-3mm}
    \captionsetup{width=.98\linewidth}
    \captionsetup{singlelinecheck = false, justification=justified}
    \caption{Our model, \OursAcronym{}, shows GAN frameworks can also be scaled up for general text-to-image synthesis tasks, generating a 512px output at an interactive speed of 0.13s, and 4096px at 3.7s. Selected examples at 2K or 4K resolutions are shown. Please zoom in for more details.  See Appendix~\ref{sec:C} and our \href{https://mingukkang.github.io/GigaGAN/}{website} for more uncurated comparisons.
    }
    \label{fig:text2image_results}
\end{figure*}

\begin{figure*}[!ht]
    \centering
    \vspace{-1mm}
    \includegraphics[width=0.97\linewidth]{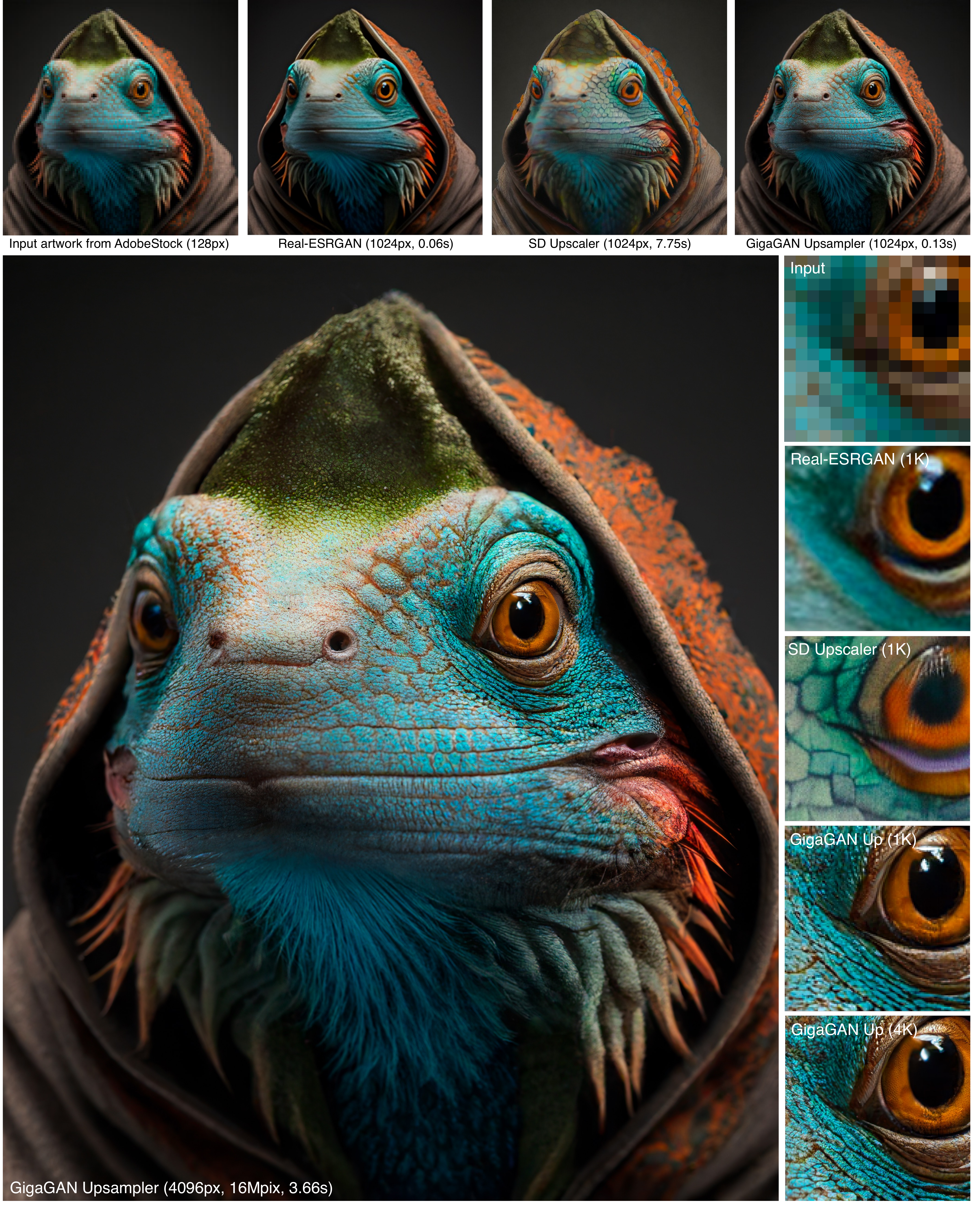}
    \vspace{-2mm}
    \caption{{\bf Our GAN-based upsampler} can serve in the upsampling pipeline of many text-to-image models that often generate initial outputs at low resolutions like 64px or 128px. We simulate such usage by applying our text-conditioned 8$\times$ superresolution model on a low-res 128px artwork to obtain the 1K output, using ``Portrait of a colored iguana dressed in a hoodie". Then our model can be re-applied to go beyond 4K. We compare our model with the text-conditioned upscaler of Stable Diffusion~\cite{stablediffusion} and unconditional Real-ESRGAN~\cite{wang2021realesrgan}. Zooming in is recommended for comparison between 1K and 4K.
    }
    \vspace{-4mm}
    \label{fig:superres}
\end{figure*}

\begin{figure*}[!ht]
    \centering
    \vspace{-1mm}
    \includegraphics[width=0.97\linewidth]{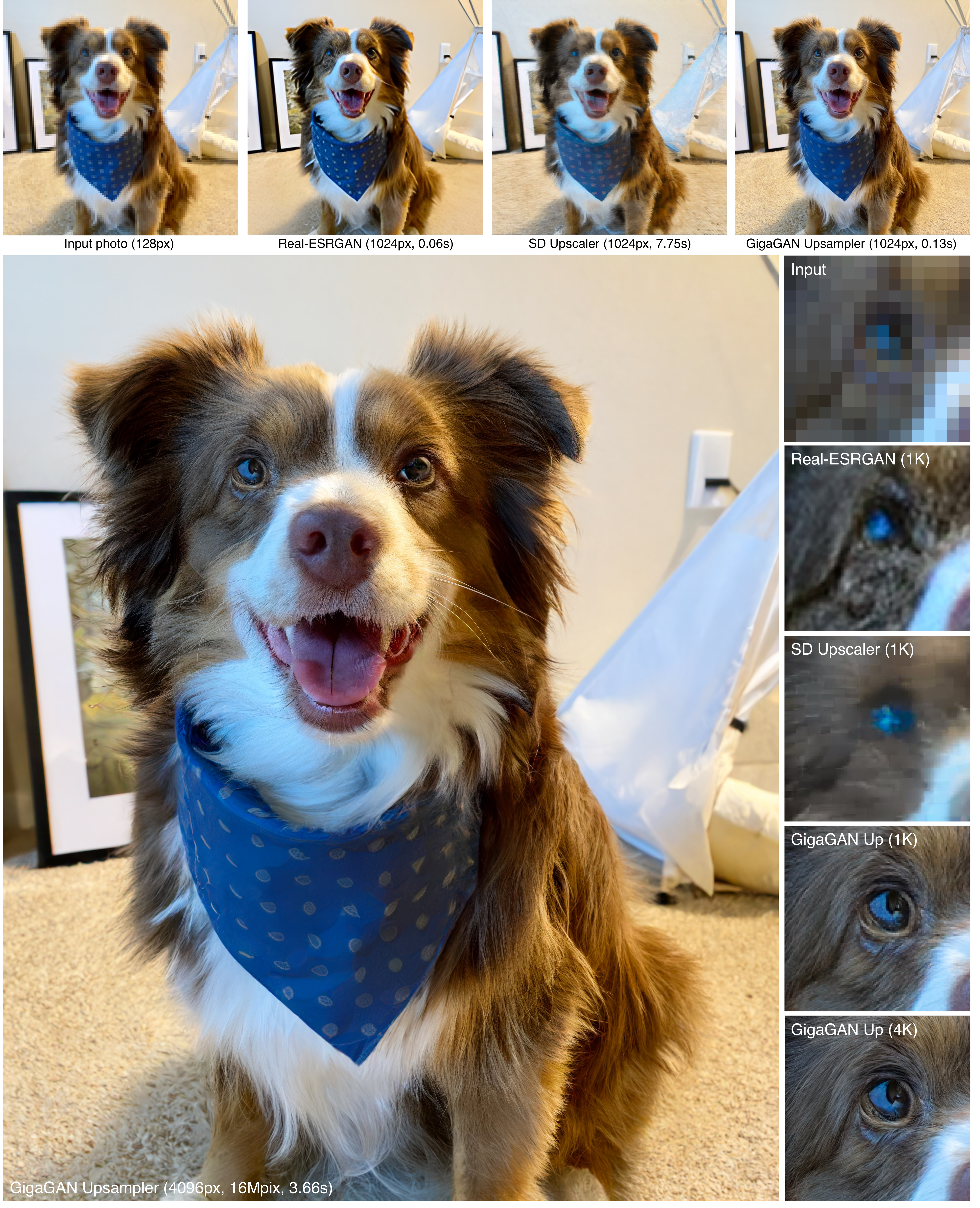}
    \vspace{-3mm}
    \caption{{\bf Our GAN-based upsampler}, similar to Figure~\ref{fig:superres}, can also be used as an off-the-shelf superresolution model for real images with a large scaling factor by providing an appropriate description of the image. We apply our text-conditioned 8$\times$ superresolution model on a low-res 128px photo to obtain the 1K output, using ``A dog sitting in front of a mini tipi tent". Then our model can be re-applied to go beyond 4K. We compare our model with the text-conditioned upscaler of Stable Diffusion~\cite{stablediffusion} and unconditional Real-ESRGAN~\cite{wang2021realesrgan}. Zooming in is recommended for comparison between 1K and 4K.}
    \vspace{-3mm}
    \label{fig:superres_dog}
\end{figure*}

Recently released models, such as DALL$\cdot$E 2~\cite{ramesh2022hierarchical}, Imagen~\cite{saharia2022photorealistic}, Parti~\cite{yu2022scaling}, and Stable Diffusion~\cite{rombach2022high}, have ushered in a new era of image generation, achieving unprecedented levels of image quality and model flexibility. 
The now-dominant paradigms, diffusion models and autoregressive models, both rely on iterative inference. This is a double-edged sword, as iterative methods enable stable training with simple objectives but incur a high computational cost during inference.

Contrast this with Generative Adversarial Networks (GANs)~\cite{Goodfellow2014GenerativeAN,radford2015unsupervised,karras2019style,Brock2019LargeSG}, which generate images through a single forward pass and thus inherently efficient. While such models dominated the previous ``era'' of generative modeling, scaling them requires careful tuning of the network architectures and training considerations due to instabilities in the training procedure. As such,
GANs have excelled at modeling single or multiple object classes, but scaling to complex datasets, much less an open world, has remained challenging. 
As a result, ultra-large models, data, and compute resources are now dedicated to diffusion and autoregressive models.
In this work, we ask -- \emph{can GANs continue to be scaled up and potentially benefit from such resources, or have they plateaued? What prevents them from further scaling, and can we overcome these barriers?}
 
We first experiment with StyleGAN2~\cite{karras2020analyzing} and observe that simply scaling the backbone causes unstable training. 
We identify several key issues and propose techniques to stabilize the training while increasing the model capacity. First, we effectively scale the generator's capacity by retaining a bank of filters and taking a sample-specific linear combination. We also adapt several techniques commonly used in the diffusion context and confirm that they bring similar benefits to GANs. For instance,  interleaving both self-attention (image-only) and cross-attention (image-text) with the convolutional layers improves performance. 

Furthermore, we reintroduce multi-scale training, finding a new scheme that improves image-text alignment and low-frequency details of generated outputs. Multi-scale training allows the GAN-based generator to use parameters in low-resolution blocks more effectively, leading to better image-text alignment and image quality. After careful tuning, we achieve stable and scalable training of a one-billion-parameter GAN (\OursAcronym{}) on large-scale datasets, such as LAION2B-en~\cite{schuhmann2022laion}. Our results are shown in Figure~\ref{fig:text2image_results}.

In addition, our method uses a multi-stage approach~\cite{denton2015deep,zhang2017stackgan}.
We first generate at $64\times 64$ and then upsample to $512\times 512$. These two networks are modular and robust enough to be used in a plug-and-play fashion. We show that our text-conditioned GAN-based upsampling network can be used as an efficient, higher-quality upsampler for a base diffusion model such as DALL$\cdot$E 2, despite never having seen diffusion images at training time (Figures~\ref{fig:superres}). 

Together, these advances enable our \OursAcronym{} to go far beyond previous GANs: 36$\times$ larger than StyleGAN2~\cite{karras2020analyzing} and 6$\times$ larger than StyleGAN-XL~\cite{sauer2022styleganxl} and XMC-GAN~\cite{zhang2021cross}. While our 1B parameter count is still lower than the largest recent synthesis models, such as Imagen (3.0B), DALL$\cdot$E 2 (5.5B), and  Parti (20B), we have not yet observed a quality saturation regarding the model size. GigaGAN achieves a zero-shot FID of 9.09 on COCO2014 dataset, lower than the FID of DALL$\cdot$E 2, Parti-750M, and Stable Diffusion.

\begin{figure*}[!ht]
    \centering
    \includegraphics[width=1.0\linewidth]{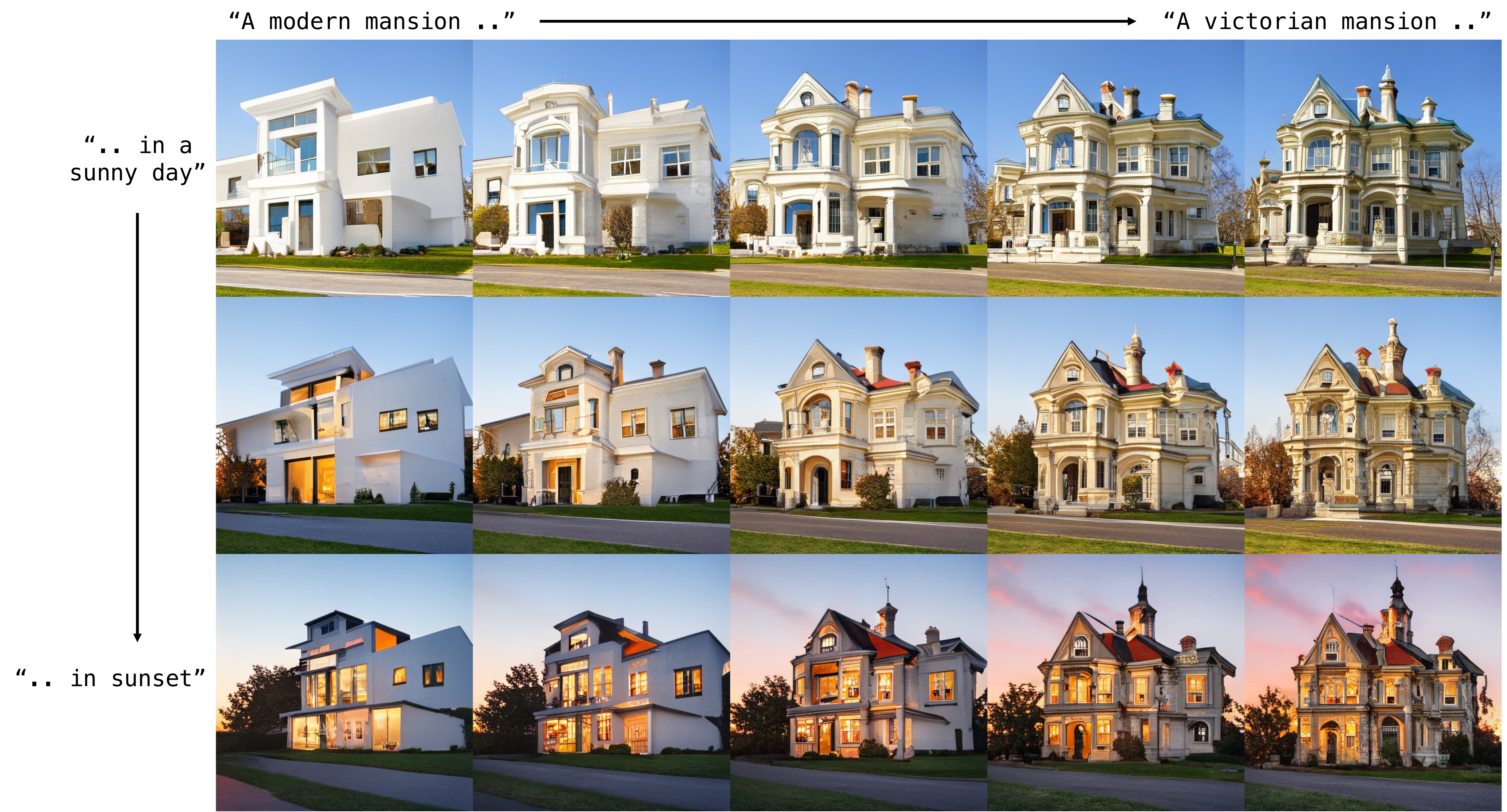}
    \vspace{-5mm}
    \caption{{\bf Prompt interpolation}. GigaGAN enables smooth interpolation between prompts, as shown in the interpolation grid. The four corners are generated from the same latent $\vb{z}$ but with different text prompts. The corresponding text embeddings $\vb{t}$ and style vectors $\vb{w}$ are interpolated to create a smooth transition. The same $\vb{z}$ results in similar layouts.  See Figure~\ref{fig:style_prompt_swapping} for more precise control.}
    \label{fig:text_interpolation}
\end{figure*}

Furthermore, GigaGAN has three major practical advantages compared to diffusion and autoregressive models. First, it is orders of magnitude faster, generating a 512px image in 0.13 seconds~(Figure~\ref{fig:text2image_results}). Second, it can synthesize ultra high-res images at 4k resolution in 3.66 seconds. Third, it is endowed with a controllable, latent vector space that lends itself to well-studied controllable image synthesis applications, such as style mixing~(Figure~\ref{fig:style_swapping}), prompt interpolation~(Figure~\ref{fig:text_interpolation}), and prompt mixing~(Figure~\ref{fig:style_prompt_swapping}). 

In summary, our model is the first GAN-based method that successfully trains a billion-scale model on billions of real-world complex Internet images. This suggests that GANs are still a viable option for text-to-image synthesis and should be considered for future aggressive scaling. Please visit our \href{https://mingukkang.github.io/GigaGAN/}{website} for additional results.

\section{Related Works}
\vspace{5pt}

\myparagraph{Text-to-image synthesis.} Generating a realistic image given a text description, explored by early works~\cite{zhu2007text,mansimov2015generating}, is a challenging task. A common approach is text-conditional GANs~\cite{reed2016learning,reed2016generative,zhang2017stackgan,xu2018attngan, zhu2019dm, tao2022df} on specific domains~\cite{wah2011caltech} 
and datasets with a closed-world assumption~\cite{lin2014microsoft}.  
With the development of diffusion models~\cite{ho2020denoisingDP, dhariwal2021diffusion}, autoregressive~(AR) transformers~\cite{chen2020generative}, and large-scale language encoders~\cite{2020t5, radford2021learning}, text-to-image synthesis has shown remarkable improvement on an open-world of arbitrary text descriptions. GLIDE~\cite{Nichol2022GLIDETP}, DALL$\cdot$E~2~\cite{ramesh2022hierarchical}, and Imagen~\cite{saharia2022photorealistic} are representative diffusion models that show photorealistic outputs with the aid of a pretrained language encoder~\cite{radford2021learning,2020t5}. 
AR models, such as DALL$\cdot$E~\cite{ramesh2021zero}, Make-A-Scene~\cite{Gafni2022MakeASceneST}, CogView~\cite{ding2021cogview,ding2022cogview2}, and Parti~\cite{yu2022scaling} also achieve amazing results. While these models exhibit unprecedented image synthesis ability, they require time-consuming iterative processes to achieve high-quality image sampling. 

To accelerate the sampling, several methods propose to reduce the sampling steps~\cite{meng2022distillation,salimans2022progressive,song2021denoising,lu2022dpm} or reuse pre-computed features~\cite{li2022efficient}. 
Latent Diffusion Model (LDM)~\cite{rombach2022high} performs the reverse processes in low-dimensional latent space instead of pixel space. However, consecutive reverse processes are still computationally expensive, limiting the usage of large-scale text-to-image models for interactive applications. 

\myparagraph{GAN-based image synthesis.} GANs~\cite{Goodfellow2014GenerativeAN} have been one of the primary families of generative models for natural image synthesis. 
As the sampling quality and diversity of GANs improve~\cite{radford2015unsupervised,karras2018progressive, karras2019style, karras2020analyzing, karras2021alias,kumari2022ensembling,sauer2021projected}, GANs have been deployed to various computer vision and graphics applications, such as text-to-image synthesis~\cite{reed2016generative}, image-to-image translation~\cite{isola2017image,zhu2017unpaired,huang2018multimodal,lee2018diverse,park2019semantic,park2020cut}, and image editing~\cite{zhu2016generative,Brock2017NeuralPE,abdal2019image2stylegan,patashnik2021styleclip}. 
Notably, StyleGAN-family models~\cite{karras2020analyzing,karras2021alias} have shown impressive ability in image synthesis tasks for single-category domains~\cite{abdal2019image2stylegan, zhu2020improved, wulff2020improving, harkonen2020ganspace,patashnik2021styleclip}. Other works have explored class-conditional GANs~\cite{Zhang2019SelfAttentionGA, Brock2019LargeSG,  Kang2021RebootingAA, zhao2020differentiable,sauer2022styleganxl} on datasets with a fixed set of object categories.

In this paper, we change the data regimes from single- or multi-categories datasets to extremely data-rich situations. We make the first expedition toward training a large-scale GAN for text-to-image generation on a vast amount of web-crawled text and image pairs, such as LAION2B-en~\cite{schuhmann2022laion} and COYO-700M~\cite{kakaobrain2022coyo-700m}. Existing GAN-based text-to-image synthesis models~\cite{reed2016generative, zhang2017stackgan, xu2018attngan, zhu2019dm, liang2020cpgan, zhang2021cross, tao2022df} are trained on relatively small datasets, such as CUB-200~(12k training pairs), MSCOCO~(82k) and LN-OpenImages~(507k). Also, those models are evaluated on associated validation datasets, which have not been validated to perform large-scale text-image synthesis like diffusion or AR models.

Concurrent with our method, StyleGAN-T~\cite{Sauer2023ARXIV} and GALIP~\cite{tao2023galip}  share similar goals and make complementary insights to ours.

\myparagraph{Super-resolution for large-scale text-to-image models.} Large-scale models require prohibitive computational costs for both training and inference. To reduce the memory and running time, cutting-edge text-to-image models~\cite{Nichol2022GLIDETP, ramesh2022hierarchical, saharia2022photorealistic, yu2022scaling} have adopted cascaded generation processes where images are first generated at $64\times64$ resolution and upsampled to $256\times256$ and $1024\times1024$ sequentially. However, the super-resolution networks are primarily based on diffusion models, which require many iterations.  
In contrast, our low-res image generators and upsamplers are based on GANs, reducing the computational costs for both stages. 
Unlike traditional super-resolution techniques~\cite{dong2015image, ledig2017photo, wang2018esrgan, anwar2020densely} that aim to faithfully reproduce low-resolution inputs or handle image degradation like compression artifacts, our upsamplers for large-scale models serve a different purpose. They need to perform larger upsampling factors while potentially leveraging the input text prompt. 

\section{Method}
\vspace{-1mm}

\begin{figure*}[ht]
    \centering
    \includegraphics[width=\figwidth]{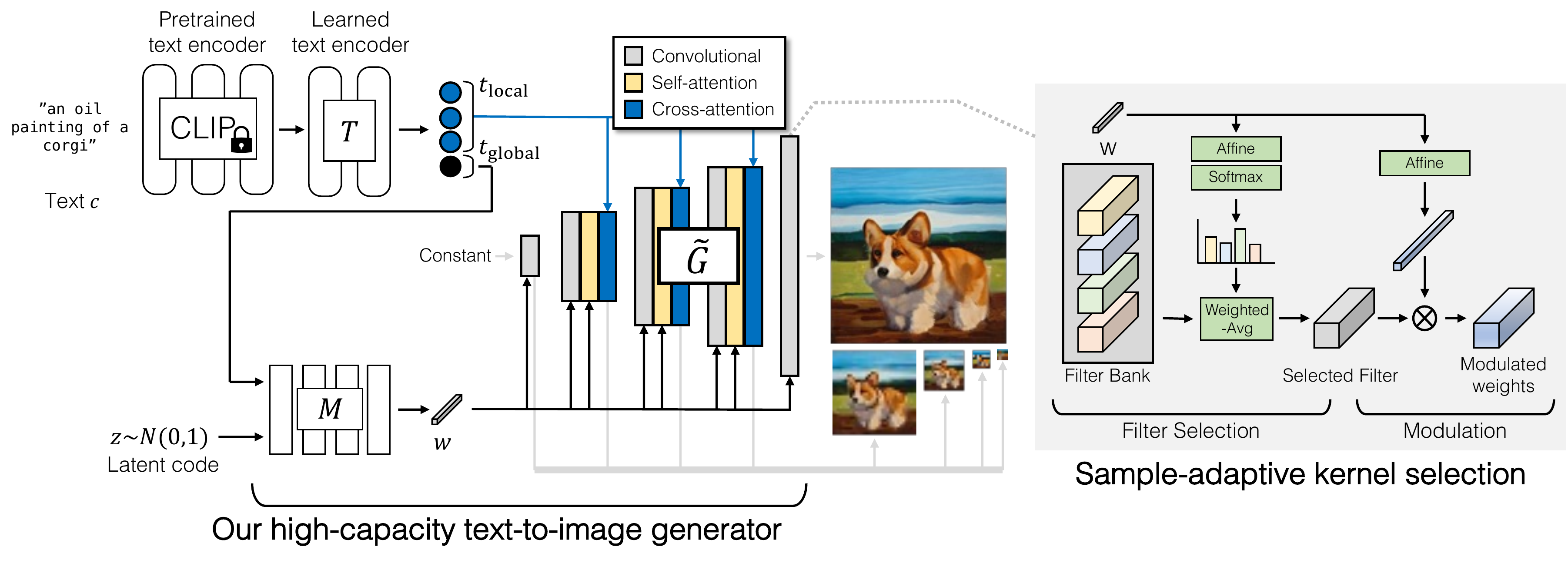}
\vspace{-4mm}
    \caption{
    {\bf Our \OursAcronym{} high-capacity text-to-image generator.} First, we extract text embeddings using a pretrained CLIP model and a learned encoder $T$. The local text descriptors are fed to the generator using cross-attention. The global text descriptor, along with a latent code $\vb{z}$, is fed to a style mapping network $M$ to produce style code $\vb{w}$. The style code modulates the main generator using our style-adaptive kernel selection, shown on the right. The generator outputs an image pyramid by converting the intermediate features into RGB images. To achieve higher capacity, we use multiple attention and convolution layers at each scale (Appendix~\ref{table:details}). We also use a separate upsampler model, which is not shown in this diagram.}
    \label{fig:pipeline}
\end{figure*}

We train a generator $G(\vb{z}, \vb{c})$ to predict an image $\vb{x}\in \mathds{R}^{H\times W\times 3}$ given a latent code $\vb{z} \sim \mathcal{N}(0, 1) \in \mathds{R}^{128}$ and text-conditioning signal $\vb{c}$. We use a discriminator $D(\vb{x}, \vb{c})$ to judge the realism of the generated image, as compared to a sample from the training database $\mathcal{D}$, which contains image-text pairs.

Although GANs~\cite{karras2018progressive,karras2019style,Brock2019LargeSG} can successfully generate realistic images on single- and multi-category datasets~\cite{karras2019style,yu2015lsun, deng2009imagenet}, 
 open-ended text-conditioned synthesis on Internet images remains challenging. 
We hypothesize that the current limitation stems from its reliance on convolutional layers. That is, the same convolution filters are challenged to model the general image synthesis function for all text conditioning across all locations of the image. In this light, we seek to inject more expressivity into our parameterization by dynamically selecting convolution filters based on the input conditioning and by capturing long-range dependence via the attention mechanism. 

Below, we discuss our key contributions to making ConvNets more expressive (\refsec{core}), followed by our designs for the generator (\refsec{generator}) and discriminator (\refsec{discriminator}). Lastly, we introduce a new, fast GAN-based upsampler model that can improve the inference quality and speed of our method and diffusion models such as Imagen~\cite{saharia2022photorealistic} and DALL$\cdot$E~2~\cite{ramesh2022hierarchical}. 

\subsection{Modeling complex contextual interaction}
\lblsec{core}

\vspace{8pt}

\myparagraph{Baseline StyleGAN generator.} We base our architecture off the conditional version of StyleGAN2~\cite{karras2020analyzing}, comprised of two networks $G = \widetilde{G}\circ M$. The mapping network $\vb{w}=M(\vb{z},\vb{c})$ maps the inputs into a ``style'' vector $\vb{w}$, which modulates a series of upsampling convolutional layers in the synthesis network $\widetilde{G}(\vb{w})$ to map a learned constant tensor to an output image $\vb{x}$. Convolution is the main engine to generate all output pixels, with the $\vb{w}$ vector as the only source of information to model conditioning.

\myparagraph{Sample-adaptive kernel selection.} To handle the highly diverse distribution of internet images, we aim to increase the capacity of convolution kernels. However, increasing the width of the convolution layers becomes too demanding, as the same operation is repeated across all locations.

We propose an efficient way to enhance the expressivity of convolutional kernels by creating them on-the-fly based on the text conditioning, as illustrated in Figure~\ref{fig:pipeline} (right). In this scheme, we instantiate a bank of $N$ filters $\{\vb{K}_i \in \mathds{R}^{C_{\text{in}} \times C_{\text{out}} \times K \times K} \}_{i=1}^N$, instead of one, that takes a feature $\vb{f} \in \mathds{R}^{C_\text{in}}$ at each layer. The style vector $\vb{w} \in \mathds{R}^{d}$ then goes through an affine layer $[W_\text{filter}, b_\text{filter}] \in \mathds{R}^{(d + 1) \times N}$ to predict a set of weights to average across the filters, to produce an aggregated filter $\vb{K} \in \mathds{R}^{C_{\text{in}} \times C_{\text{out}} \times K \times K} $. 

\begin{equation}
\label{eq:adaptive_kernel_selection}
    \vb{K} = \sum_{i=1}^N \vb{K}_i \cdot \text{softmax}\big( W_\text{filter}^{\top}\vb{w} + b_\text{filter}\big)_i
\end{equation}

\noindent The filter is then used in the regular convolution pipeline of StyleGAN2, with the second affine layer $[W_\text{mod}, b_\text{mod}] \in \mathds{R}^{(d + 1) \times C_\text{in}}$ for weight (de-)modulation~\cite{karras2020analyzing}. 

\begin{equation}
    g_\text{adaconv}(\vb{f}, \vb{w}) = \big((W_{\text{mod}}^{\top}\vb{w} + b_\text{mod} \big) \otimes \vb{K}) *\vb{f},
\end{equation}

\noindent where $\otimes$ and $*$ represent (de-)modulation and convolution.

At a high level, the softmax-based weighting can be viewed as a differentiable filter selection process based on input conditioning. Furthermore, since the filter selection process is performed only once at each layer, the selection process is much faster than the actual convolution, decoupling compute complexity from the resolution. Our method shares a spirit with dynamic convolutions~\cite{jia2016dynamic,wu2018pay,ha2016hypernetworks,tanjim2020dynamicrec} in that the convolution filters dynamically change per sample, but differs in that we explicitly instantiate a larger filter bank and select weights based on a separate pathway conditional on the $\vb{w}$-space of StyleGAN. 

\myparagraph{Interleaving attention with convolution.} 
Since the convolutional filter operates within its receptive field, it cannot contextualize itself in relationship to distant parts of the images. One way to incorporate such long-range relationships is using attention layers $g_\text{attention}$. While recent diffusion-based models~\cite{dhariwal2021diffusion, Ho2022CascadedDM, rombach2022high} have commonly adopted attention mechanisms, StyleGAN architectures are predominantly convolutional with the notable exceptions such as BigGAN~\cite{Brock2019LargeSG}, GANformer~\cite{hudson2021generative}, and ViTGAN~\cite{lee2022vitgan}. 

We aim to improve the performance of StyleGAN by integrating attention layers with the convolutional backbone. However, simply adding attention layers to StyleGAN often results in training collapse, possibly because the dot-product self-attention is not Lipschitz, as pointed out by Kim et al.~\cite{kim2021lipschitz}. As the Lipschitz continuity of discriminators has played a critical role in stable training~\cite{arjovsky2017WGAN,gulrajani2017WGANGP,mescheder2018R1}, we use the L2-distance instead of the dot product as the attention logits to promote Lipschitz continuity~\cite{kim2021lipschitz}, similar to ViTGAN~\cite{lee2022vitgan}. 

To further improve performance, we find it crucial to match the architectural details of StyleGAN, such as equalized learning rate~\cite{karras2018progressive} and weight initialization from a unit normal distribution. We scale down the L2 distance logits to roughly match the unit normal distribution at initialization and reduce the residual gain from the attention layers. We further improve stability by tying the key and query matrix~\cite{lee2022vitgan}, and applying weight decay.

In the synthesis network $\widetilde{G}$, the attention layers are interleaved with each convolutional block, leveraging the style vector $\vb{w}$ as an additional token. At each attention block, we add a separate cross-attention mechanism $g_\text{cross-attention}$ to attend to individual word embeddings~\cite{bahdanau2015neural}. We use each input feature tensor as the query, and the text embeddings as the key and value of the attention mechanism.

\subsection{Generator design}
\lblsec{generator}
\vspace{5pt}

\myparagraph{Text and latent-code conditioning.} First, we extract the text embedding from the prompt. Previous works~\cite{ramesh2021zero,saharia2022photorealistic} have shown that leveraging a strong language model is essential for producing strong results. To do so, we tokenize the input prompt (after padding it to $C=77$ words, following best practices~\cite{ramesh2021zero,saharia2022photorealistic}) to produce conditioning vector $\vb{c}\in \mathds{R}^{C\times 768}$, and take the features from the penultimate layer~\cite{saharia2022photorealistic} of a frozen CLIP feature extractor~\cite{radford2021learning}. To allow for additional flexibility, we apply additional attention layers $T$ on top to process the word embeddings before passing them to the MLP-based mapping network. This results in text embedding $\vb{t}=T(\mathcal{E}_{\text{txt}}(\vb{c})) \in \mathds{R}^{C\times 768}$. Each component $\vb{t}_i$ of $\vb{t}$ captures the embedding of the $i^{\text{th}}$ word in the sentence. We refer to them as $\vb{t}_\text{local} = \vb{t}_{\{1:C\} \setminus \text{EOT}} \in \mathds{R}^{(C - 1)\times 768}$. The EOT (``end of text") component of $\vb{t}$ aggregates global information, and is called $\vb{t}_\text{global} \in \mathds{R}^{768}$. We process this global text descriptor, along with the latent code $\vb{z} \sim \mathcal{N}(0, 1)$, via an MLP mapping network to extract the style $\vb{w} = M(\vb{z},\vb{t}_\text{global})$. 
\begin{equation}
\begin{split}
    (\vb{t}_\text{local}&, \vb{t}_\text{global}) = T(\mathcal{E}_{\text{txt}}(\vb{c})), \\ 
    &\vb{w} = M(\vb{z},\vb{t}_\text{global}).
\end{split}
\end{equation}
\noindent 
Different from the original StyleGAN, we use both the text-based style code $\vb{w}$ to modulate the synthesis network $\widetilde{G}$ and the word embeddings $\vb{t}_\text{local}$ as features for cross-attention.

\begin{equation}
    \vb{x} = \widetilde{G}(\vb{w}, \vb{t}_\text{local}).
\end{equation}

Similar to earlier works~\cite{mansimov2015generating, saharia2022photorealistic,ramesh2022hierarchical}, the text-image alignment visually improves with cross-attention. 

\myparagraph{Synthesis network.} Our synthesis network consists of a series of upsampling convolutional layers, with each layer enhanced with the adaptive kernel selection (Equation~\ref{eq:adaptive_kernel_selection}) and followed by our attention layers. 

\begin{equation}
    \hspace{-1ex}\vb{f}_{\ell+1}=g^\ell_\text{xa}(g^\ell_\text{attn}(g^\ell_\text{adaconv}(\vb{f}_\ell, \vb{w}), \vb{w}), \vb{t}_\text{local}), 
\end{equation}

\noindent where $g^\ell_\text{xa}$, $g^\ell_\text{attn}$, and $g^\ell_\text{adaconv}$ denote the $l$-th layer of cross-attention, self-attention, and weight (de-)modulation layers. We find it beneficial to increase the depth of the network by adding more blocks at each layer. In addition, our generator outputs a multi-scale image pyramid with $L=5$ levels, instead of a single image at the highest resolution, similar to MSG-GAN~\cite{karnewar2020msg} and AnycostGAN~\cite{lin2021anycost}. We refer to the pyramid as $\{\vb{x}_i\}_{i=0}^{L-1} = \{\vb{x}_0, \vb{x}_1, ..., \vb{x}_4\}$, with spatial resolutions $\{S_i\}_{i=0}^{L-1} = \{64, 32, 16, 8, 4\}$, respectively. The base level $\vb{x}_0$ is the output image $\vb{x}$. Each image of the pyramid is independently used to compute the GAN loss, as discussed in Section~\ref{sec:discriminator}.
We follow the findings of StyleGAN-XL~\cite{sauer2022styleganxl} and turn off the style mixing and path length regularization~\cite{karras2020analyzing}. We include more training details in Appendix~\ref{sec:text2img}.

\subsection{Discriminator design}
\lblsec{discriminator}
\vspace{5pt}

\begin{figure}[!t]
    \centering
    \includegraphics[width=\figwidth]{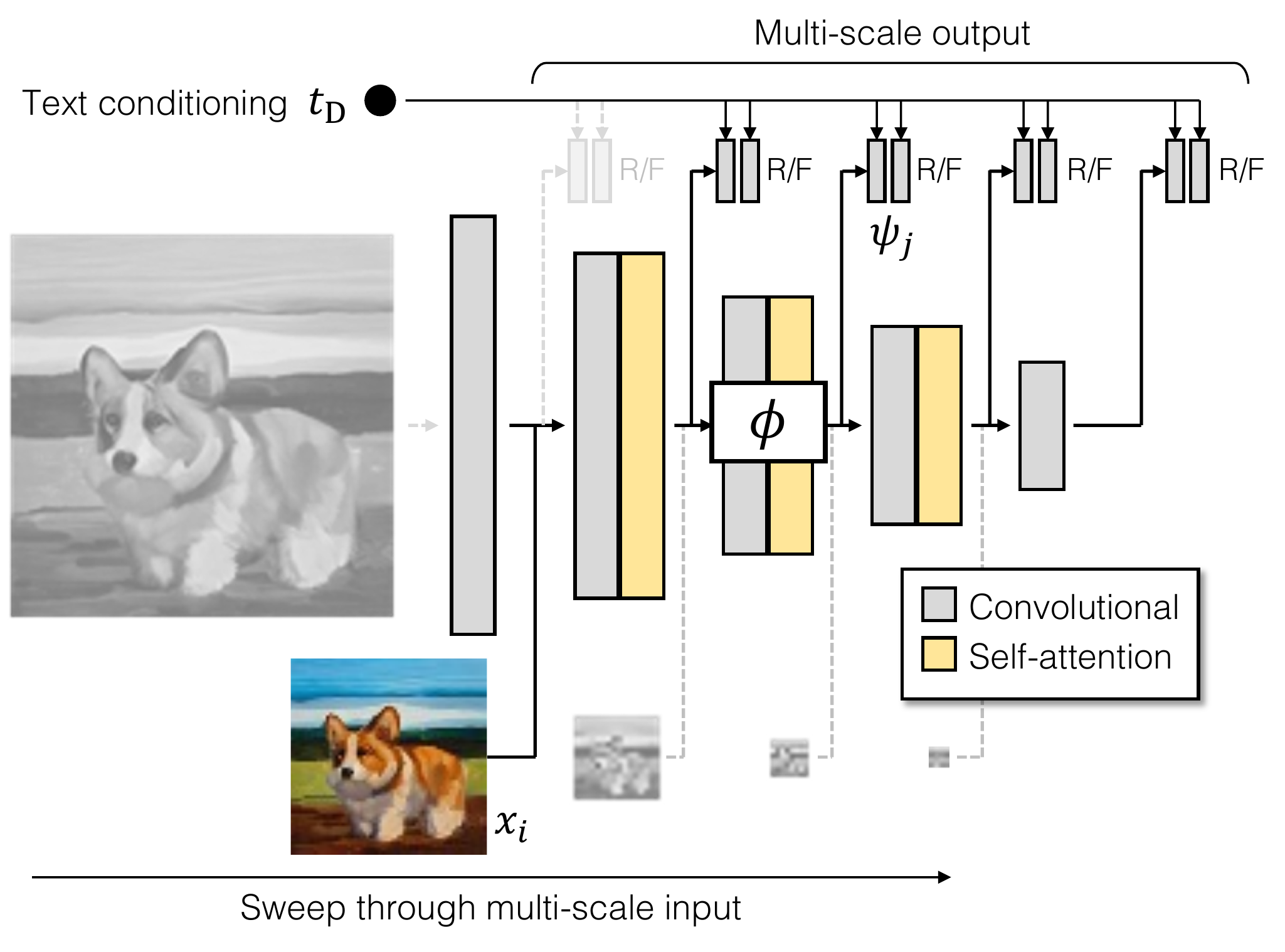}
\vspace{-0mm}
    \caption{{\bf Our discriminator} consists of two branches for processing the image and the text conditioning $t_D$. The text branch processes the text similar to the generator (Figure~\ref{fig:pipeline}). The image branch receives an image pyramid and makes independent predictions for each image scale. Moreover, the predictions are made at all subsequent scales of the downsampling layers, making it a \textit{multi-scale input, multi-scale output} (MS-I/O) discriminator.}
    
    \label{fig:D_architecture}
    \vspace{-0mm}
\end{figure}

As shown in \reffig{D_architecture}, our discriminator consists of separate branches for processing text with the function $\vb{t}_D$ and images with function $\phi$. The prediction of real vs. fake is made by comparing the features from the two branches using function $\psi$. We introduce a new way of making predictions on multiple scales. Finally, we use additional CLIP and Vision-Aided GAN losses~\cite{kumari2022ensembling} to improve stability.

\myparagraph{Text conditioning.} First, to incorporate conditioning into discriminators, we extract text descriptor $\vb{t}_\text{D}$ from text $\vb{c}$. Similar to the generator, we apply a pretrained text encoder, such as CLIP~\cite{radford2021learning}, followed by a few learnable attention layers. In this case, we only use the global descriptor.

\myparagraph{Multiscale image processing.}
We observe that the early, low-resolution layers of the generator become inactive, using small dynamic ranges irrespective of the provided prompts. StyleGAN2~\cite{karras2020analyzing} also observes this phenomenon, concluding that the network relies on the high-resolution layers, as the model size increases. As recovering performance in low frequencies, which contains complex structure information, is crucial, we redesign the model architecture to provide training signals across multiple scales.

Recall the generator produces a pyramid $\{\vb{x}_i\}_{i=0}^{L-1}$, with the full image $\vb{x}_0$ at the pyramid base. MSG-GAN~\cite{karnewar2020msg} improves performance by making a prediction on the entire pyramid at once, enforcing consistency across scales. 
However, in our large-scale setting, this harms stability, as this limits the generator from making adjustments to its initial low-res output.

Instead, we process each level of the pyramid \textit{independently}. As shown in \reffig{D_architecture}, each level $\vb{x}_i$ makes a real/fake prediction at multiple scales $i < j \leq L$. 
For example, the full $\vb{x}_0$ makes predictions at $L=5$ scales, the next level $\vb{x}_1$ makes predictions at 4 scales, and so on. In total, our discriminator produces $\frac{L(L+1)}{2}$ predictions, supervising multi-scale generations at multiple scales.

To extract features at different scales, we define feature extractor $\phi_{i\rightarrow j}: \mathds{R}^{X_i\times X_i\times 3} \rightarrow \mathds{R}^{X_j^D\times X_j^D\times C_j}$. Practically, each sub-network $\phi_{i\rightarrow j}$ is a subset of full $\phi\triangleq \phi_{0\rightarrow L}$, with $i>0$ indicating late entry and $j<L$ indicating early exit. Each layer in $\phi$ is composed of self-attention, followed by convolution with stride 2. The final layer flattens the spatial extent into a $1\times 1$ tensor. This produces output resolutions at $\{X_j^D\} = \{32, 16, 8, 4, 1\}$. 
This allows us to inject lower-resolution images on the pyramid into intermediate layers~\cite{karras2018progressive}. As we use a shared feature extractor across different levels and most of the added predictions are made at low resolutions, the increased computation overhead is manageable.

\myparagraph{Multi-scale input, multi-scale output adversarial loss.} In total, our training objective consists of discriminator losses, along with our proposed matching loss, to encourage the discriminator to take into account the conditioning:
\vspace{-8pt}
\begin{equation}
\begin{split}
    \mathcal{V}_\text{MS-I/O}(G, D) = \sum_{i=0}^{L-1} \sum_{j=i+1}^{L} \hspace{1mm} &\mathcal{V}_\text{GAN}(G_i, D_{ij}) + \mathcal{V}_\text{match}(G_i, D_{ij}),
    \label{eq:GAN}
\end{split}
\end{equation}

\noindent where $\mathcal{V}_\text{GAN}$ is the standard, non-saturating GAN loss~\cite{Goodfellow2014GenerativeAN}.

\noindent To compute the discriminator output, we train predictor $\psi$, which uses text feature $\vb{t}_\text{D}$ to modulate image features $\phi(\vb{x})$:

\begin{equation}
\begin{split}
    D_{ij}(\vb{x}, \vb{c}) &= \psi_j(\phi_{i\rightarrow j}(\vb{x}_i), \vb{t}_D) + \text{Conv}_\text{1$\times$1}(\phi_{i\rightarrow j}(\vb{x}_i)),
    \label{eq:SPD}
\end{split}
\end{equation}

\noindent where $\psi_j$ is implemented as a 4-layer $1\times 1$ modulated convolution, and $\text{Conv}_{1\times 1}$ is added as a skip connection to explicitly maintain an unconditional prediction branch~\cite{Miyato2018cGANsWP}.

\myparagraph{Matching-aware loss.}
The previous GAN terms measure how closely the image $\vb{x}$ matches the conditioning $\vb{c}$, as well as how realistic $\vb{x}$ looks, irrespective of conditioning. However, during early training, when artifacts are obvious, the discriminator heavily relies on making a decision independent of conditioning and hesitates to account for the conditioning.

To enforce the discriminator to incorporate conditioning, we match $\vb{x}$ with a random, independently sampled condition $\hat{\vb{c}}$, and present them as a fake pair: 

\begin{equation}
\begin{split}
\mathcal{V_{\text{match}}} = 
{\mathbb{E}_{\vb{x},\vb{c},\hat{\vb{c}}}} \hspace{.5mm} \big[ \log_{} &(1 + \exp( D(\vb{x}, \hat{\vb{c}}))) \\ 
  + &\log_{} (1 + \exp( D(G(\vb{c}), \hat{\vb{c}})) \big],
\end{split}
    \label{eq:match}
\end{equation}

\noindent where $(\vb{x}, c)$ and $\hat{\vb{c}}$ are separately sampled from $p_{\text{data}}$. This loss has previously been explored in text-to-image GAN works~\cite{reed2016generative,zhang2017stackgan}, except we find that enforcing the Matching-aware loss on generated images from $G$, as well real images $\vb{x}$, leads to clear gains in performance (Table~\ref{tab:ablation}).

\myparagraph{CLIP contrastive loss.} We further leverage off-the-shelf pretrained models as a loss function~\cite{sauer2021projected,kumari2022ensembling,sungatullina2018image}. In particular, we enforce the generator to produce outputs that are identifiable by the pre-trained CLIP image and text encoders~\cite{radford2021learning}, $\mathcal{E}_\text{img}$ and $\mathcal{E}_\text{txt}$, in the contrastive cross-entropy loss that was used to train them originally. 

\begin{align*}
    \mathcal{L}_\text{CLIP} =
    {\mathbb{E}_{\{\vb{c}_n \} }} \hspace{.2mm} \Big[ - \log \frac{
    \text{exp}(\mathcal{E}_{\text{img}}(G(\vb{c}_0))^\top \mathcal{E}_{\text{txt}}(\vb{c}_0) )}{
    \sum_{n}{\text{exp} (\mathcal{E}_{\text{img}}(G( \vb{c}_0))^\top \mathcal{E}_{\text{txt}}(\vb{c}_n)  }})
    \Big],
    \tag{9}
    \label{eq:clip_loss}
\end{align*}
\noindent 
where $\{ \vb{c}_n \} = \{\vb{c}_0, \dots \}$ are sampled captions from the training data.

\myparagraph{Vision-aided adversarial loss.} Lastly, we build an additional discriminator that uses the CLIP model as a backbone, known as Vision-Aided GAN~\cite{kumari2022ensembling}. We freeze the CLIP image encoder, extract features from the intermediate layers, and process them through a simple network with $3\times 3$ conv layers to make real/fake predictions. We also incorporate conditioning through modulation, as in Equation~\ref{eq:SPD}. 
To stabilize training, we also add a fixed random projection layer, as proposed by Projected GAN~\cite{sauer2021projected}. We refer to this as $\mathcal{L}_\text{Vision}(G)$ (omitting the learnable discriminator parameters for clarity).

Our final objective is $\mathcal{V}(G, D) = \mathcal{V}_\text{MS-I/O}(G,D) + \mathcal{L}_\text{CLIP}(G) + \mathcal{L}_\text{Vision}(G)$, with weighting between the terms specified in Table~\ref{table:details}.

\subsection{GAN-based upsampler}
\label{sec:upsampler}
\vspace{-1mm}
Furthermore, GigaGAN framework can be easily extended to train a text-conditioned superresolution model, capable of upsampling the outputs of the base GigaGAN generator to obtain high-resolution images at 512px or 2k resolution. By training our pipeline in two separate stages, we can afford a higher capacity 64px base model within the same computational resources. 

In the upsampler, the synthesis network is rearranged to an asymmetric U-Net architecture, which processes the 64px input through 3 downsampling residual blocks, followed by 6 upsampling residual blocks with attention layers to produce the 512px image. There exist skip connections at the same resolution, similar to CoModGAN~\cite{zhao2021comodgan}. The model is trained with the same losses as the base model, as well as the LPIPS Perceptual Loss~\cite{zhang2018unreasonable} with respect to the ground truth high-resolution image. Vision-aided GAN is not used for the upsampler. During training and inference time, we apply moderate Gaussian noise augmentation to reduce the gap between real and GAN-generated images. Please refer to Appendix~\ref{sec:super-res} for more details. 

Our GigaGAN framework becomes particularly effective for the superresolution task compared to the diffusion-based models, which cannot afford as many sampling steps as the base model at high resolution. The LPIPS regression loss also provides a stable learning signal. We believe that our GAN upsampler can serve as a drop-in replacement for the superresolution stage of other generative models.

\section{Experiments}
\vspace{-1mm}

\begin{figure*}[!ht]
    \centering
    \includegraphics[width=1.0\linewidth]{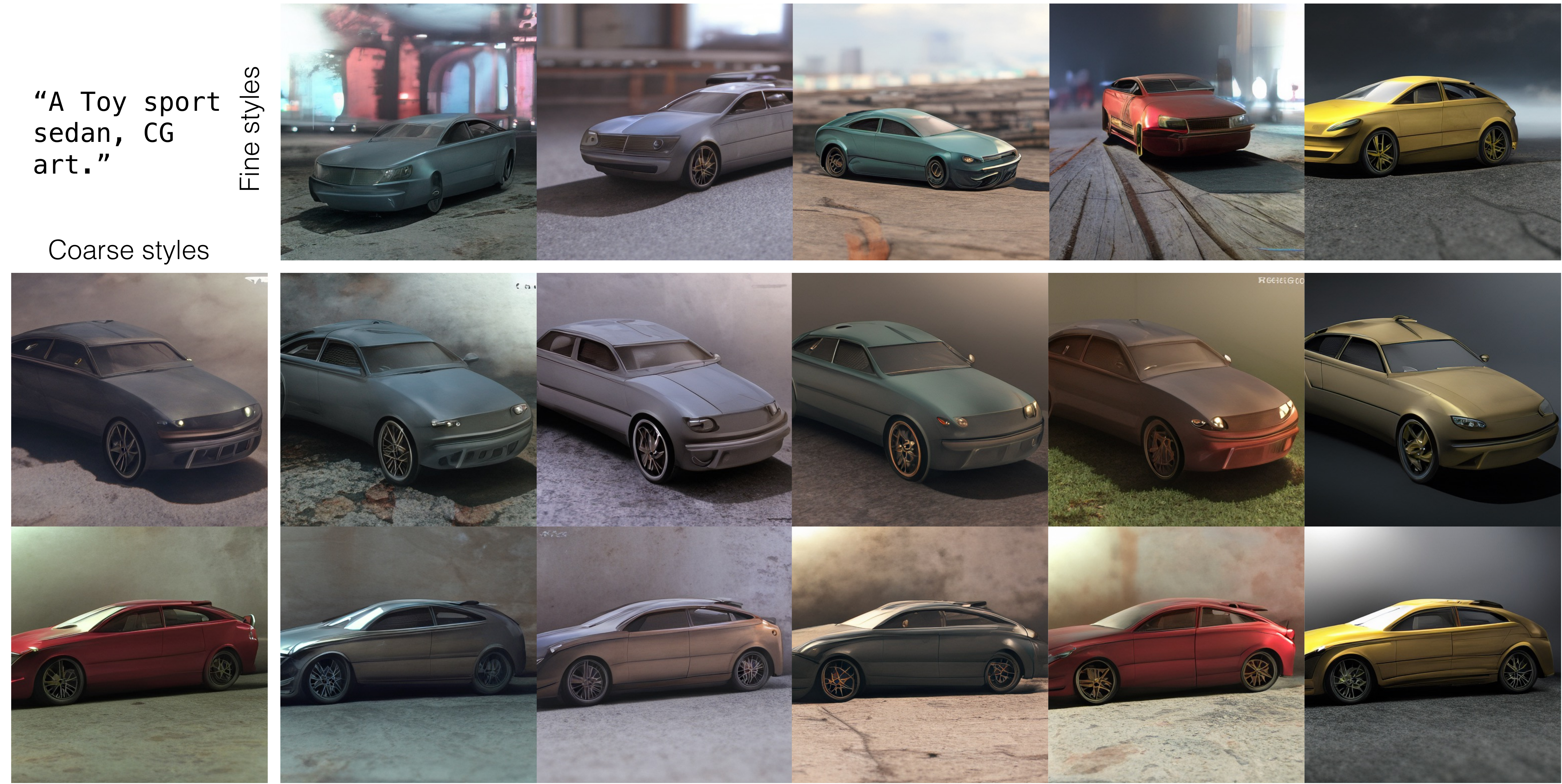}
    \vspace{-5mm}
    \caption{{\bf Style mixing}. Our GAN-based architecture retains a disentangled latent space, enabling us to blend the coarse style of one sample with the fine style of another. All outputs are generated with the prompt ``A Toy sport sedan, CG art." The corresponding latent codes are spliced together to produce a style-swapping grid.}
    \label{fig:style_swapping}
\end{figure*}
\begin{figure*}[!ht]
    \centering
    \includegraphics[width=1.0\linewidth]{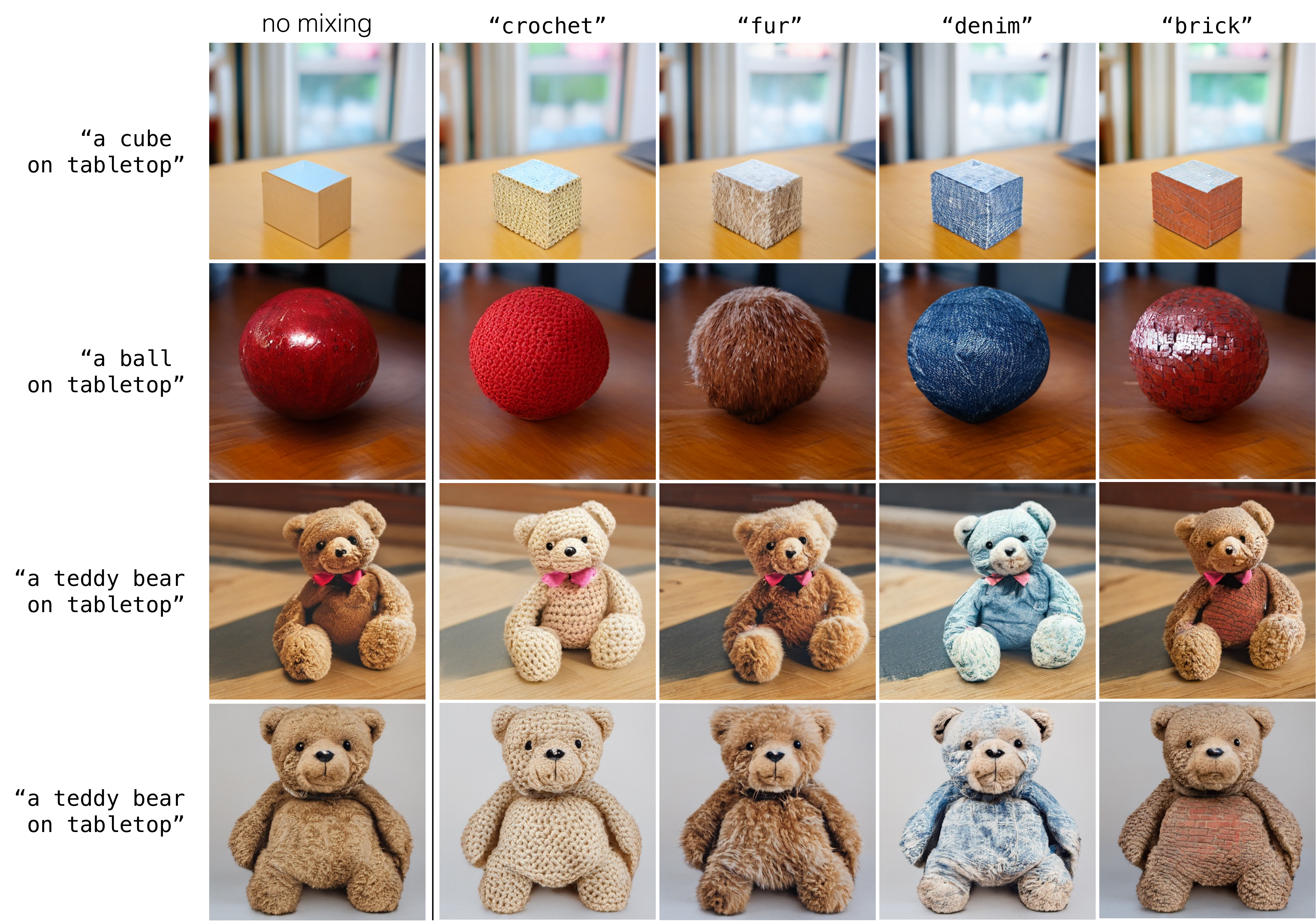}
    \vspace{-5mm}
    \caption{{\bf Prompt mixing}. GigaGAN retains a disentangled latent space, enabling us to combine the coarse style of one sample with the fine style of another. Moreover, GigaGAN can directly control the style with text prompts. Here we generate four outputs using the prompts ``a X on tabletop", shown in the ``no mixing'' column. Then we re-compute the text embeddings $\vb{t}$ and the style codes $\vb{w}$ using the new prompts ``a X with the texture of Y on tabletop", such as ``a cube with the texture of crochet on tabletop", and apply them to the second half layers of the generator, achieving layout-preserving fine style control. Cross-attention mechanism automatically localizes the style to the object of interest.}
    
    \label{fig:style_prompt_swapping}
\end{figure*}

Systematic, controlled evaluation of large-scale text-to-image synthesis tasks is difficult, as most existing models are not publicly available. Training a new model from scratch would be prohibitively costly, even if the training code were available. Still, we compare our model to recent text-to-image models, such as Imagen~\cite{saharia2022photorealistic}, Latent Diffusion Models~(LDM)~\cite{rombach2022high}, Stable Diffusion~\cite{stablediffusion}, and Parti~\cite{yu2022scaling}, based on the available information, while acknowledging considerable differences in the training dataset, number of iterations, batch size, and model size. 
In addition to text-to-image results, we evaluate our model on ImageNet class-conditional generation in Appendix~\ref{sec:B}, for an apples-to-apples comparison with other methods at a more controlled setting.

For quantitative evaluation, we mainly use the \fid (FID)~\cite{Heusel2017GANsTB} for measuring the realism of the output distribution and the CLIP score for evaluating the image-text alignment. 

We conduct five different experiments. First, we show the effectiveness of our method by gradually incorporating each technical component one by one (Section~\ref{sec:ablation}). Second, our text-to-image synthesis results demonstrate that GigaGAN exhibits comparable FID with Stable Diffusion (SD-v1.5)~\cite{rombach2022high} while generating results hundreds of times faster than diffusion or autoregressive models (Section~\ref{sec:text2img_main}). Third, we compare GigaGAN with a distillation-based diffusion model~\cite{meng2022distillation} and show that GigaGAN can synthesize higher-quality images faster than the distillation-based diffusion model. Fourth, we verify the advantage of GigaGAN's upsampler over other upsamplers in both conditional and unconditional super-resolution tasks. Lastly, we show our large-scale GANs still enjoy the continuous and disentangled latent space manipulation of GANs, enabling new image editing modes (\refsec{editing_expr}).

\subsection{Training and evaluation details}
\vspace{-1mm}
We implement \OursAcronym{} based on the StudioGAN PyTorch library~\cite{kang2022studiogan}, following the standard FID evaluation protocol with the anti-aliasing bicubic resize function~\cite{parmar2021cleanfid}, unless otherwise noted. For text-to-image synthesis, we train our models on the union of LAION2B-en~\cite{schuhmann2022laion} and COYO-700M~\cite{kakaobrain2022coyo-700m} datasets, with the exception of the 128-to-1024 upsampler model trained on Adobe's internal Stock images. The image-text pairs are preprocessed based on CLIP score~\cite{hessel2021clipscore}, image resolution, and aesthetic score~\cite{laionaesthetic}, similar to prior work~\cite{stablediffusion}. We use CLIP ViT-L/14~\cite{radford2021learning} for the pre-trained text encoder and OpenCLIP ViT-G/14~\cite{ilharco_gabriel_2021_5143773} for CLIP score calculation~\cite{hessel2021clipscore} except for Table 1. All our models are trained and evaluated on A100 GPUs. We include more training and evaluation details in Appendix~\ref{sec:A}.

\subsection{Effectiveness of proposed components}
\lblsec{ablation}
\vspace{-1mm}

\begin{table}[t]
\caption{\textbf{Ablation study on 64px text-to-image synthesis.} To evaluate the effectiveness of our components, we start with a modified version of StyleGAN for text conditioning. While increasing the network width does not show satisfactory improvement, each addition of our contributions keeps improving metrics.  Finally, we increase the network width and scale up training to reach our final model. All ablated models are trained for 100k steps at a batch size of 256 except for the Scale-up row (1350k iterations with a larger batch size). CLIP Score is computed using CLIP ViT-B/32~\cite{radford2021learning}.}
\vspace{-2mm}
\resizebox{0.47\textwidth}{!}{
\begin{tabular}{lccc}
\cmidrule[1.0pt]{1-4}
Model                     & FID-10k~$\downarrow$ & CLIP Score~$\uparrow$ & \# Param.\\
\cmidrule[1.0pt]{1-4}
StyleGAN2 &  29.91 & 0.222 & 27.8M \\
~~~~~+~Larger (5.7$\times$)  & 34.07 & 0.223 & 158.9M \\
\cmidrule[1.0pt]{1-4}
~~+~Tuned & 28.11 & 0.228 &  26.2M \\
~~+~Attention      &  23.87  & 0.235 & 59.0M \\
~~~~~+~Matching-aware D & 27.29 & 0.250& 59.0M\\
~~+~Matching-aware G and D & 21.66 &  0.254 & 59.0M\\
~~+~Adaptive convolution  & 19.97 & 0.261 & 80.2M\\
~~+~Deeper         &  19.18 & 0.263  & 161.9M \\
~~+~CLIP loss & 14.88 &  0.280 &  161.9M \\
~~+~Multi-scale training & 14.92 & 0.300 & 164.0M\\
~~+~Vision-aided GAN & 13.67 & 0.287 &  164.0M \\
\cmidrule[1.0pt]{1-4}
~~+~Scale-up (\textbf{GigaGAN}) & 9.18 & 0.307 & 652.5M \\
\cmidrule[1.0pt]{1-4}
\end{tabular}
}
\vspace{-1mm}
\label{tab:ablation}
\end{table}

First, we show the effectiveness of our formulation via ablation study in Table~\ref{tab:ablation}. We set up a baseline by adding text-conditioning to StyleGAN2 and tuning the configuration based on the findings of StyleGAN-XL. 
We first directly increase the model size of this baseline, but we find that this does not improve the FID and CLIP scores. Then, we add our components one by one and observe that they consistently improve performance. In particular, our model is more scalable, as the higher-capacity version of the final formulation achieves better performance. 

\subsection{Text-to-Image synthesis}
\lblsec{text2img_main}
We proceed to train a larger model by increasing the capacity of the base generator and upsampler to 652.5M and 359.1M, respectively. This results in an unprecedented size of GAN model, with a total parameter count of 1.0B. Table~\ref{tab:t2i_results} compares the performance of our end-to-end pipeline to various text-to-image generative models~\cite{ramesh2021zero,Nichol2022GLIDETP,rombach2022high,ramesh2022hierarchical,saharia2022photorealistic,balaji2022ediffi,yu2022scaling,zhou2021lafite,stablediffusion,chang2023muse}. Note that there exist differences in the training dataset, the pretrained text encoders, and even image resolutions. For example, GigaGAN initially synthesizes 512px images, which are resized to 256px before evaluation. 

Table~\ref{tab:t2i_results} shows that GigaGAN exhibits a lower FID than DALL$\cdot$E 2~\cite{ramesh2022hierarchical}, Stable Diffusion~\cite{stablediffusion}, and Parti-750M~\cite{yu2022scaling}. While our model can be optimized to better match the feature distribution of real images than existing models,  the quality of the generated images is not necessarily better (see Appendix~\ref{sec:C} for more samples). We acknowledge that this may represent a corner case of zero-shot FID on COCO2014 dataset and suggest that further research on a better evaluation metric is necessary to improve text-to-image models. Nonetheless, we emphasize that GigaGAN is the first GAN model capable of synthesizing promising images from arbitrary text prompts and exhibits competitive zero-shot FID with other text-to-image models.

\begin{table}[t]
\caption{{\bf Comparison to recent text-to-image models}. Model size, total images seen during training, COCO FID-30k, and inference speed of text-image models. ${*}$ denotes that the model has been evaluated by us.  GigaGAN achieves a lower FID than DALL$\cdot$E 2~\cite{ramesh2022hierarchical}, Stable Diffusion~\cite{stablediffusion}, and Parti-750M~\cite{yu2022scaling}, while being much faster than competitive methods. GigaGAN and SD-v1.5 require 4,783 and 6,250 A100 GPU days, and Imagen and Parti need approximately 4,755 and 320 TPUv4 days for training.}

\vspace{-3.5mm}

\resizebox{0.47\textwidth}{!}{
\begin{tabular}{llcccccc}
\cmidrule[1.0pt]{1-7}
 & Model & Type &  \# Param. & \# Images & FID-30k~$\downarrow$ & Inf. time\\
\cmidrule[1.0pt]{1-7}
\parbox[t]{0mm}{\multirow{10}{*}{\rotatebox[origin=c]{90}{256}}}  & GLIDE~\cite{Nichol2022GLIDETP} & Diff & 5.0B & 5.94B & 12.24 & 15.0s  \\
& LDM~\cite{rombach2022high} & Diff & 1.5B & 0.27B & 12.63 & 9.4s  \\
& DALL$\cdot$E 2~\cite{ramesh2022hierarchical} & Diff & 5.5B & 5.63B & 10.39 & -  \\
& Imagen~\cite{saharia2022photorealistic} & Diff & 3.0B & 15.36B & 7.27 & 9.1s  \\
& eDiff-I~\cite{balaji2022ediffi} & Diff & 9.1B & 11.47B & 6.95 & 32.0s  \\
& DALL$\cdot$E~\cite{ramesh2021zero} & AR  & 12.0B & 1.54B & 27.50 & -  \\
& Parti-750M~\cite{yu2022scaling} & AR & 750M & 3.69B & 10.71 & -  \\
& Parti-3B~\cite{yu2022scaling} & AR & 3.0B & 3.69B & 8.10 & 6.4s  \\
& Parti-20B~\cite{yu2022scaling} & AR & 20.0B & 3.69B & 7.23 & - \\
& LAFITE~\cite{zhou2021lafite} & GAN & 75M & - & 26.94 & 0.02s\\
\cmidrule[1.0pt]{1-7}
\parbox[t]{0mm}{\multirow{3}{*}{\rotatebox[origin=c]{90}{512}}}
& SD-v1.5$^{*}$~\cite{stablediffusion} & Diff & 0.9B & 3.16B & 9.62 & 2.9s \\
& Muse-3B~\cite{chang2023muse} & AR & 3.0B & 0.51B & 7.88 & 1.3s \\
& \textbf{GigaGAN} & GAN & 1.0B & 0.98B & 9.09 & 0.13s\\
\cmidrule[1.0pt]{1-7}
\end{tabular}}
\label{tab:t2i_results}
\vspace{-15pt}
\end{table}

\subsection{Comparison with distilled diffusion models}
\lblsec{sd_distilled}
While GigaGAN is at least 20 times faster than the above diffusion models, there have been efforts to improve the inference speed of diffusion models. We compare GigaGAN with progressively distilled Stable Diffusion~(SD-distilled)~\cite{meng2022distillation}. Table~\ref{tab:sd_distileed} demonstrates that GigaGAN remains faster than the distilled Stable Diffusion while showing better FID and CLIP scores of 21.1 and 0.32, respectively. We follow the evaluation protocol of SD-distilled~\cite{meng2022distillation} and report FID and CLIP scores on COCO2017 dataset~\cite{lin2014microsoft}, where images are resized to 512px.

\subsection{Super-resolution for large-scale image synthesis}

\lblsec{supres_expr}
\vspace{-1mm}
\begin{table}[t]

\caption{ { \bf Comparison to distilled diffusion models} shows that GigaGAN achieves better FID and CLIP scores compared to the progressively distilled diffusion models~\cite{meng2022distillation} for fast inference. As GigaGAN generates outputs in a single feedforward pass, the inference speed is still faster. The evaluation setup is different from Table~\ref{tab:t2i_results} to match SD-distilled's protocol~\cite{meng2022distillation}. }
\vspace{-3mm}
\resizebox{0.45\textwidth}{!}{
\begin{tabular}{lcccc}
\cmidrule[1.0pt]{1-5}

Model    & Steps & FID-5k~$\downarrow$ & CLIP~$\uparrow$ & Inf.~time\\
\cmidrule[1.0pt]{1-5}
SD-distilled-2~\cite{meng2022distillation}  &  2 & 37.3 & 0.27 &  0.23s\\
SD-distilled-4~\cite{meng2022distillation}  &  4 & 26.0 & 0.30 & 0.33s\\
SD-distilled-8~\cite{meng2022distillation}  &  8 & 26.9 & 0.30 & 0.52s\\
SD-distilled-16~\cite{meng2022distillation}  & 16 & 28.8 & 0.30 & 0.88s \\
\cmidrule[1.0pt]{1-5}
\textbf{GigaGAN}     & 1 & 21.1 & 0.32 & 0.13s\\
\cmidrule[1.0pt]{1-5}
\label{tab:sd_distileed}
\end{tabular}}
\vspace{-5.5mm}
\end{table}
\begin{table}[t]
\caption{ { \bf Text-conditioned 128$\rightarrow$1024 super-resolution} on random 10K LAION samples, compared against unconditional Real-ESRGAN~\cite{wang2021realesrgan} and Stable Diffusion Upscaler~\cite{stablediffusion}. GigaGAN enjoys the fast speed of a GAN-based model while achieving better FID, patch-FID~\cite{chai2022anyresolution}, CLIP score, and LPIPS~\cite{zhang2018unreasonable}.}
\vspace{-3mm}
\resizebox{0.48\textwidth}{!}{
\begin{tabular}{lcccccc}
\cmidrule[1.0pt]{1-7}
Model & \# Param. & Inf. time & FID-10k~$\downarrow$ & pFID~$\downarrow$& CLIP~$\uparrow$ & LPIPS$\downarrow$ \\
\cmidrule[1.0pt]{1-7}
Real-ESRGAN~\cite{wang2021realesrgan}  & 17M & 0.06s & 8.60 & 22.8 & 0.314& 0.363\\
SD Upscaler~\cite{stablediffusion} & 846M & 7.75s & 9.39& 41.3& 0.316& 0.523\\
\cmidrule[1.0pt]{1-7}
\textbf{GigaGAN}         & 693M & 0.13s & 1.54 & 8.90 & 0.322 & 0.274\\
\cmidrule[1.0pt]{1-7}
\end{tabular}}
\label{table:superres_cond}
\end{table}

\begin{table}[t]
\caption{ { \bf Unconditional 64$\rightarrow$256 super-resolution} on ImageNet. We compare to a simple U-Net trained with a pixel regression loss (U-Net regression), and diffusion-based methods (SR3~\cite{saharia2022image} and LDM~\cite{rombach2022high}. Our method achieves higher realism scores represented by the Inception Score (IS) and FID.}
\vspace{-3mm}
\resizebox{0.48\textwidth}{!}{
\begin{tabular}{lcccccc}
\cmidrule[1.0pt]{1-7}
Model            & \# Param. & Steps & IS~$\uparrow$ & FID-50k~$\downarrow$ & PSNR~$\uparrow$ & SSIM~$\uparrow$ \\
\cmidrule[1.0pt]{1-7}
U-Net regression~\cite{saharia2022image} & 625M & 1 &121.1 & 15.2 & 27.9 & 0.80 \\
SR3~\cite{saharia2022image}              & 625M & 100 &180.1 & 5.2 & 26.4 & 0.76 \\
LDM-4~\cite{rombach2022high}            & 169M & 100 &166.3 & 2.8 & 24.4 & 0.69 \\
emphLDM-4~\cite{rombach2022high}        & 552M & 100 &174.9 & 2.4 & 24.7 & 0.71 \\
LDM-4-G~\cite{rombach2022high}            & 183M & 50 & 153.7 & 4.4 & 25.8 & 0.74 \\
\cmidrule[1.0pt]{1-7}
\textbf{GigaGAN}                                    & 359M & 1 & 191.5  &  1.2   &  24.3    &  0.71    \\
\cmidrule[1.0pt]{1-7}
\end{tabular}}
\label{table:superres}
\vspace{-5mm}
\end{table}
We separately evaluate the performance of the GigaGAN upsampler. Our evaluation consists of two parts. First, we compare GigaGAN with several commonly-used upsamplers. 
For the text-conditioned upsampling task, we combine the Stable Diffusion~\cite{stablediffusion} 4x Upscaler and 2x Latent Upscaler to establish an 8x upscaling model (SD Upscaler). We also use the unconditional Real-ESRGAN~\cite{wang2021realesrgan} as another baseline. Table~\ref{table:superres_cond} measures the performance of the upsampler on random 10K images from the LAION dataset and shows that our GigaGAN upsampler significantly outperforms the other upsamplers in realism scores (FID and patch-FID~\cite{chai2022anyresolution}), text alignment (CLIP score) and closeness to the ground truth (LPIPS~\cite{zhang2018unreasonable}). In addition, for more controlled comparison, we train our model on the ImageNet {\em unconditional} superresolution task and compare performance with the diffusion-based models, including SR3~\cite{saharia2022image} and LDM~\cite{rombach2022high}. As shown in Table~\ref{table:superres}, GigaGAN achieves the best IS and FID scores with a single feedforward pass.

\subsection{Controllable image synthesis}
\lblsec{editing_expr}
\vspace{-1mm}
StyleGANs are known to possess a linear latent space useful for image manipulation, called the $\mathcal{W}$-space. Likewise, we perform coarse and fine-grained style swapping using style vectors $\vb{w}$. Similar to the $\mathcal{W}$-space of StyleGAN, Figure~\ref{fig:style_swapping} illustrates that \OursAcronym{} maintains a disentangled $\mathcal{W}$-space, suggesting existing latent manipulation techniques of StyleGAN can transfer to \OursAcronym{}. Furthermore, our model possesses another latent space of text embedding $\vb{t} = [\vb{t}_\text{local}, \vb{t}_\text{global}]$ prior to $\mathcal{W}$, and we explore its potential for image synthesis. 
In Figure~\ref{fig:style_prompt_swapping}, we show that the disentangled style manipulation can be controlled via text inputs. In detail, we can compute the text embedding $\vb{t}$ and style code $\vb{w}$ using different prompts and apply them to different layers of the generator. This way, we gain not only the coarse and fine style disentanglement but also an intuitive prompt-based maneuver in the style space.

\vspace{-1mm}
\section{Discussion and Limitations}
\vspace{-1mm}

\begin{figure}[t]
    \centering
    \includegraphics[width=\figwidth]{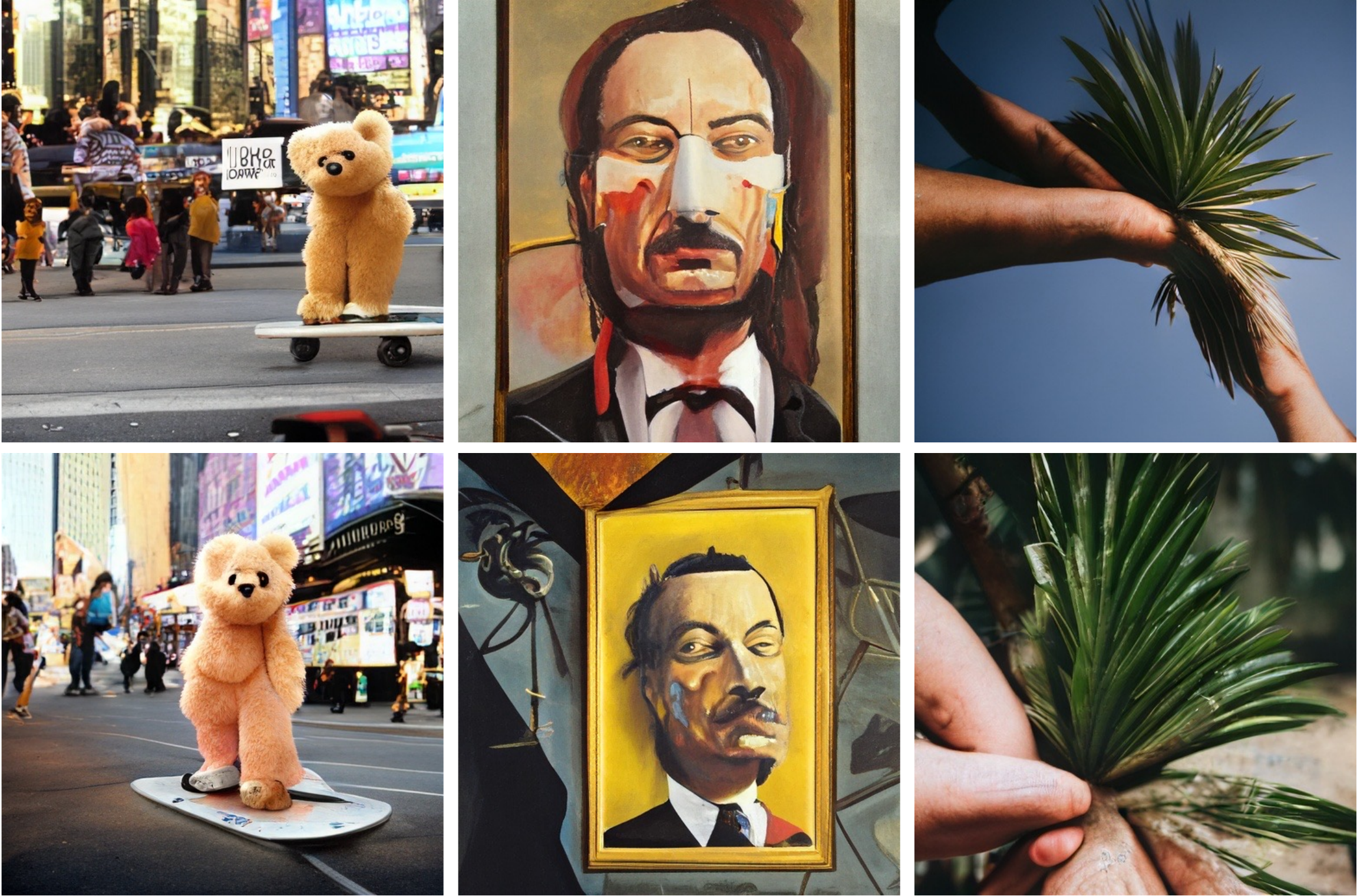}
\vspace{-3mm}
    \caption{{\bf Failure cases.} Our outputs with the same prompts as DALL$\cdot$E~2. Each column conditions on ``a teddy bear on a skateboard in Times Square", ``a Vibrant portrait painting of Salvador Dali with a robotic half face", and ``A close up of a handpalm with leaves growing from it". Compared to production-grade models such as DALL$\cdot$E~2, our model exhibits limitations in realism and compositionality. See Appendix~\ref{sec:C} for uncurated comparisons.}
    
    \label{fig:limitation}
    \vspace{-6mm}
\end{figure}

Our experiments provide a conclusive answer about the scalability of GANs: our new architecture can scale up to model sizes that enable text-to-image synthesis.  However, 
the visual quality of our results is not yet comparable to production-grade models like DALL$\cdot$E~2. Figure~\ref{fig:limitation} shows several instances where our method fails to produce high-quality results when compared to DALL$\cdot$E~2, in terms of photorealism and text-to-image alignment for the same input prompts used in their paper.

Nevertheless, we have tested capacities well beyond what is possible with a na\"ive approach and achieved competitive results with autoregressive and diffusion models trained with similar resources while being orders of magnitude faster and enabling latent interpolation and stylization. Our \OursAcronym{} architecture opens up a whole new design space for large-scale generative models and brings back key editing capabilities that became challenging with the transition to autoregressive and diffusion models.  We expect our performance to improve with larger models, as seen in Table~\ref{tab:ablation}.

\myparagraph{Acknowledgments.} We thank Simon Niklaus, Alexandru Chiculita, and Markus Woodson for building the distributed training pipeline.  We thank Nupur Kumari, Gaurav Parmar, Bill Peebles, Phillip Isola, Alyosha Efros, and Joonghyuk Shin for their helpful comments. We also want to thank Chenlin Meng, Chitwan Saharia, and Jiahui Yu for answering many questions about their fantastic work. We thank Kevin Duarte for discussions regarding upsampling beyond 4K. Part of this work was done while Minguk Kang was an intern at Adobe Research. Minguk Kang and Jaesik Park were supported by IITP grant funded by the government of South Korea (MSIT) (POSTECH GSAI: 2019-0-01906 and Image restoration: 2021-0-00537).

\balance
{\small
\bibliographystyle{ieee_fullname}
\bibliography{paper}
}

\clearpage

\appendix
\section*{\Large{Appendices}}
\addcontentsline{toc}{section}{Appendices}
\renewcommand\thefigure{A\arabic{figure}}
\renewcommand{\thetable}{A\arabic{table}}
\setcounter{figure}{0}
\setcounter{table}{0}

We first provide training and evaluation details in Appendix~\ref{sec:A}. Then, we share results on ImageNet, with visual comparison to existing methods in Appendix~\ref{sec:B}. Lastly in Appendix~\ref{sec:C}, we show more visuals on our text-to-image synthesis results and compare them with LDM~\cite{rombach2022high}, Stable Diffusion~\cite{stablediffusion}, and DALL$\cdot$E 2~\cite{ramesh2022hierarchical}.
\section{Training and evaluation details}
\lblsec{A}
\vspace{8pt}

\subsection{Text-to-image synthesis}
\lblsec{text2img}
\vspace{-1mm}

We train GigaGAN on a combined dataset of LAION2B-en~\cite{schuhmann2022laion} and COYO-700M~\cite{kakaobrain2022coyo-700m} in PyTorch framework~\cite{NEURIPS2019_9015}. For training, we apply center cropping, which results in a square image whose length is the same as the shorter side of the original image. Then, we resize the image to the resolution $64\times64$ using PIL.LANCZOS~\cite{Turkowski1990FiltersFC} resizer, which supports anti-aliasing~\cite{parmar2021cleanfid}. We filter the training image--text pairs based on image resolution~($\geq512$), CLIP score~($>0.3$)~\cite{hessel2021clipscore}, aesthetics score~($>5.0$)~\cite{laionaesthetic}, and remove watermarked images. We train our GigaGAN based on the configurations denoted in the fourth and fifth columns of Table~\ref{table:details}.

For evaluation, we use 40,504 and 30,000 real and generated images from COCO2014~\cite{lin2014microsoft} validation dataset as described in  Imagen~\cite{saharia2022photorealistic}. We apply the center cropping and resize the real and generated images to $299\times299$ resolution using PIL.BICUBIC, suggested by clean-fid~\cite{parmar2021cleanfid}. We use the clean-fid library~\cite{parmar2021cleanfid} for FID calculation.

\subsection{Conditional image synthesis on ImageNet}
\lblsec{imagenet}
\vspace{-1mm}

We follow the training and evaluation protocol proposed by Kang~\etal~\cite{kang2022studiogan} to make a fair comparison against other cutting-edge generative models. We use the same cropping strategy to process images for training and evaluation as in our text-to-image experiments. Then, we resize the image to the target resolution ($64\times64$  for the base generator or $256\times256$ for the super-resolution stack) using PIL.LANCZOS~\cite{Turkowski1990FiltersFC} resizer, which supports anti-aliasing~\cite{parmar2021cleanfid}. Using the pre-processed training images, we train GigaGAN based on the configurations denoted in the second and third columns of Table~\ref{table:details}.

For evaluation, we upsample the real and generated images to $299\times299$ resolution using the PIL.BICUBIC resizer. To compute FID, we generate 50k images without truncation tricks~\cite{karras2019style, Brock2019LargeSG} and compare those images with the entire training dataset. We use the pre-calculated features of real images provided by StudioGAN~\cite{kang2022studiogan} and 50k generated images for Precision \& Recall~\cite{Kynknniemi2019ImprovedPA} calculation.

\subsection{Super-resolution results}
\lblsec{super-res}
\vspace{-1mm}

For model training, we preprocess ImageNet in the same way as in \refsec{imagenet} and use the configuration in the last column of Table~\ref{table:details}. To compare our model with SR3~\cite{saharia2022image} and LDM fairly, we follow the evaluation procedure described in SR3 and LDM papers.

\section{ImageNet experiments}
\lblsec{B}
\vspace{8pt}
\subsection{Qualitative results}
We train a class-conditional GAN on the ImageNet dataset~\cite{deng2009imagenet}, for which apples-to-apples comparison is possible using the same dataset and evaluation pipeline. Our GAN achieves comparable generation quality to the cutting-edge generative models without a pretrained ImageNet classifier, which acts favorably toward automated metrics~\cite{kynkaanniemi2022role}. We apply L2 self-attention, style-adaptive convolution kernel, and matching-aware loss to our model and use a wider synthesis network 
 to train the base 64px model with a batch size of 1024. Additionally, we train a separate 256px class-conditional upsampler model and combine them with an end-to-end finetuning stage. Table~\ref{table:imagenet256} shows that our method generates high-fidelity images. 

\begin{table}[ht!]
\caption{ {\bf Class-conditional synthesis on ImageNet 256px.} Our method performs competitively against large diffusion and transformer models. Shaded methods leverage a pretrained ImageNet classifier at training or inference time, which could act favorably toward the automated metrics~\cite{kynkaanniemi2022role}. $\dag$ indicates IS~\cite{Salimans2016ImprovedTF} and FID~\cite{Heusel2017GANsTB} are borrowed from the original DiT paper~\cite{peebles2022scalable}.}
\vspace{-1mm}
\resizebox{0.48\textwidth}{!}{
\begin{tabular}{llcccc}
\cmidrule[1.0pt]{1-6}
 & Model & IS~\cite{Salimans2016ImprovedTF} & FID~\cite{Heusel2017GANsTB} & Precision/Recall~\cite{Kynknniemi2019ImprovedPA} &  Size\\
\cmidrule[1.0pt]{1-6}
\parbox[t]{0mm}{\multirow{2}{*}{\rotatebox[origin=c]{90}{GAN}}} & BigGAN-Deep\cite{Brock2019LargeSG} & 224.46 & 6.95 & 0.89/0.38  & 112M \\
& \colorbox{lightgray}{StyleGAN-XL}\cite{sauer2022styleganxl} & \colorbox{lightgray}{297.62} & \colorbox{lightgray}{2.32} & \colorbox{lightgray}{0.82/0.61} & \colorbox{lightgray}{166M}\\
\cmidrule[0.3pt]{1-6}
\parbox[t]{0mm}{\multirow{6}{*}{\rotatebox[origin=c]{90}{Diffusion}}} & \colorbox{lightgray}{ADM-G}\cite{dhariwal2021diffusion} & \colorbox{lightgray}{207.86} & \colorbox{lightgray}{4.48} & \colorbox{lightgray}{0.84/0.62} & \colorbox{lightgray}{608M}\\
& \colorbox{lightgray}{ADM-G-U}\cite{dhariwal2021diffusion}       & \colorbox{lightgray}{240.24} & \colorbox{lightgray}{4.01} & \colorbox{lightgray}{0.85/0.62} &  \colorbox{lightgray}{726M}\\
& CDM\cite{Ho2022CascadedDM}         & 158.71 & 4.88 & - / -  & -\\
& LDM-8-G\cite{rombach2022high} & 209.52 & 7.76 & - / - &   506M\\
& LDM-4-G\cite{rombach2022high}         & 247.67 & 3.60 & - / - & 400M\\
& DiT-XL/2$^{\dag}$\cite{peebles2022scalable}         & 278.24 & 2.27 & - / - & 675M\\
\cmidrule[0.3pt]{1-6}
\parbox[t]{0mm}{\multirow{3}{*}{\rotatebox[origin=c]{90}{xformer}}} & Mask-GIT\cite{chang2022maskgit} & 216.38 & 5.40 & 0.87/0.60& 227M\\
& \colorbox{lightgray}{VQ-GAN}\cite{esser2021taming} & \colorbox{lightgray}{314.61} & \colorbox{lightgray}{5.20} & \colorbox{lightgray}{0.81/0.57} &  \colorbox{lightgray}{1.4B}\\
& \colorbox{lightgray}{RQ-Transformer}\cite{lee2022autoregressive} & \colorbox{lightgray}{339.41} & \colorbox{lightgray}{3.83} & \colorbox{lightgray}{0.85/0.60} &  \colorbox{lightgray}{3.8B}\\
\cmidrule[1.0pt]{1-6}
& \textbf{GigaGAN}        & 225.52 & 3.45 & 0.84/0.61 &  569M\\
\cmidrule[1.0pt]{1-6}
\end{tabular}}
\label{table:imagenet256}
\vspace{-2mm}
\end{table}

\subsection{Quantitative results}
\begin{figure*}[t!]
    \centering
    \includegraphics[width=0.95\linewidth]{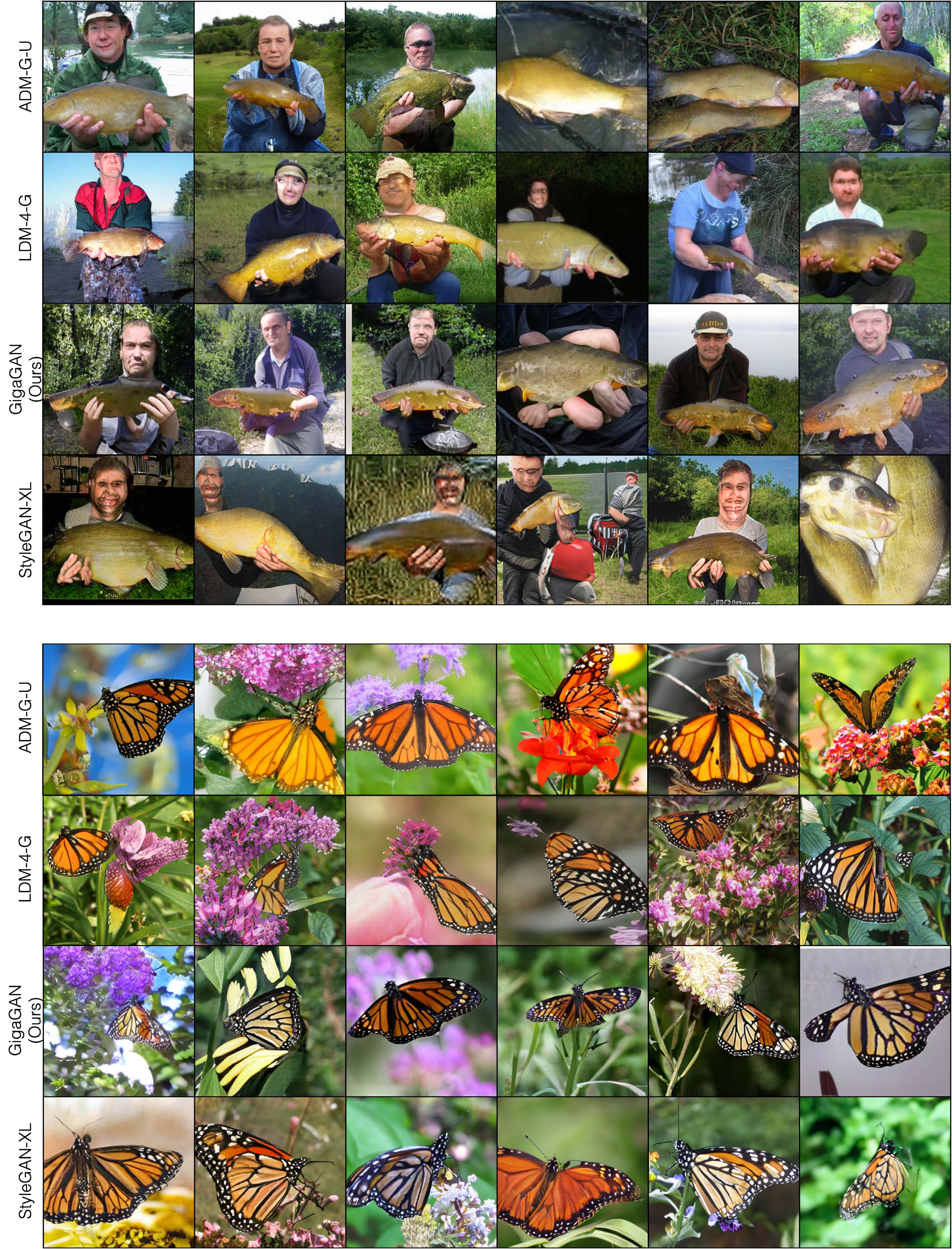}
\vspace{-3mm}
    \caption{Uncurated images (above: Tench and below: Monarch) from ADM-G-U~\cite{dhariwal2021diffusion}, LDM-4-G~\cite{rombach2022high}, GigaGAN~(ours), and StyleGAN-XL~\cite{sauer2022styleganxl}. FID values of each generative model are 4.01, 3.60, 3.45, and 2.32, respectively.}
    \label{fig:imagenet1}
    \vspace{-3mm} 
\end{figure*}

\begin{figure*}[t!]
    \centering
    \includegraphics[width=0.95\linewidth]{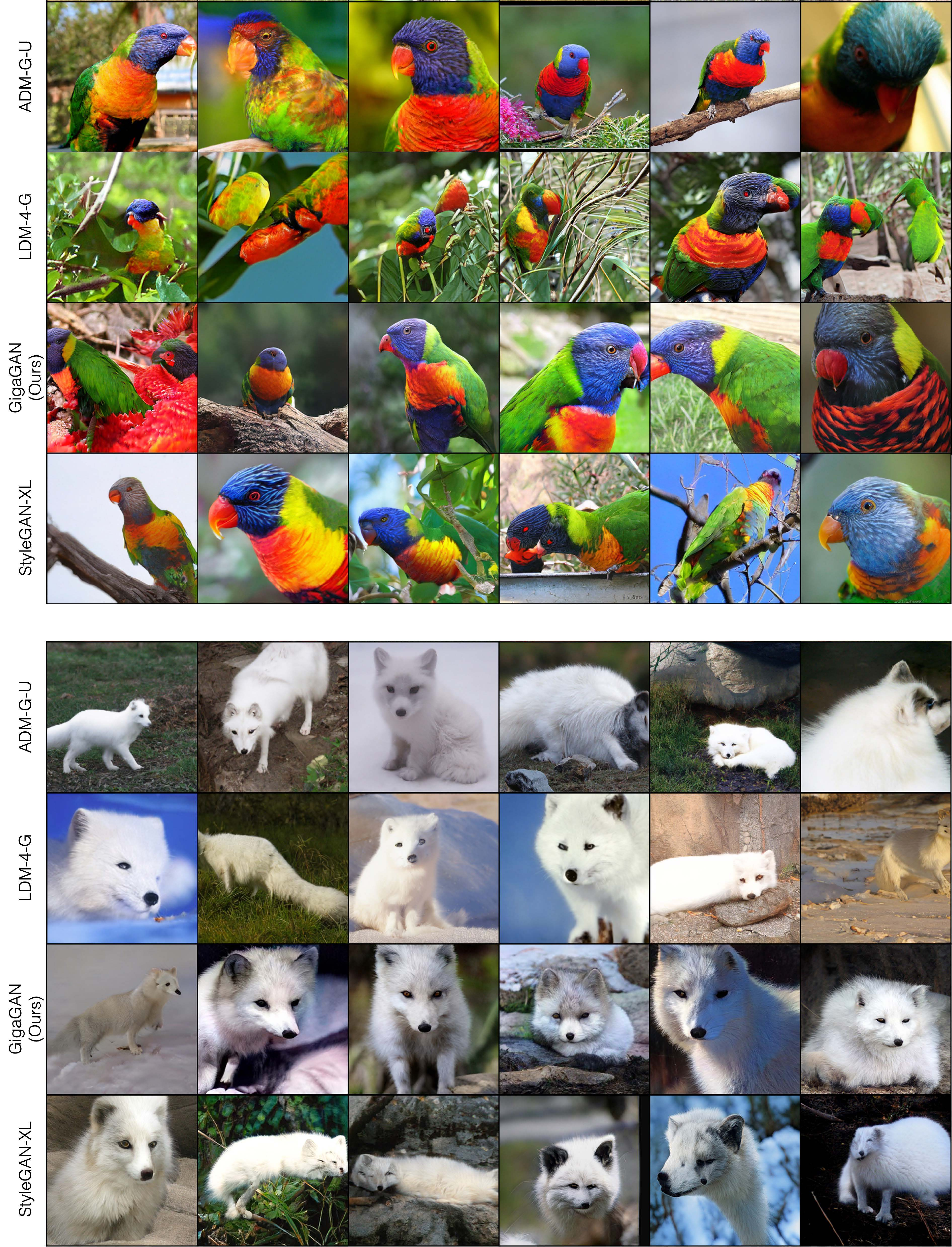}
\vspace{-3mm}
    \caption{Uncurated images (above: Lorikeet and below: Arctic fox) from ADM-G-U~\cite{dhariwal2021diffusion}, LDM-4-G~\cite{rombach2022high}, GigaGAN~(ours), and StyleGAN-XL~\cite{sauer2022styleganxl}. FID values of each generative model are 4.01, 3.60, 3.45, and 2.32, respectively.}
    \label{fig:imagenet2}
    \vspace{-3mm} 
\end{figure*}

We provide visual results from ADM-G-U, LDM, StyleGAN-XL~\cite{sauer2022styleganxl}, and GigaGAN in Figures~\ref{fig:imagenet1} and \ref{fig:imagenet2}. Although StyleGAN-XL has the lowest FID, its visual quality appears worse than ADM and GigaGAN.
StyleGAN-XL struggles to synthesize the overall image structure, leading to less realistic images.  
In contrast, GigaGAN appears to synthesize the overall structure better than StyleGAN-XL and faithfully captures fine-grained details, such as the wing patterns of a monarch and the white fur of an arctic fox. Compared to GigaGAN, ADM-G-U synthesizes the image structure more rationally but lacks in reflecting the aforementioned fine-grained details. 

\begin{center}
\begin{table*}[t!]
\caption{ {\bf Hyperparameters for GigaGAN training.}  We denote Projection Discriminator~\cite{Miyato2018cGANsWP} as PD, R1 regularization~\cite{Mescheder2018ICML} as R1,  Learned Perceptual Image Patch Similarity~\cite{zhang2018unreasonable} as LPIPS, Adam with decoupled weight decay~\cite{loshchilov2018decoupled} as AdamW, and the pretrained VIT-B/32 visual encoder~\cite{radford2021learning} as CLIP-ViT-B/32-V.}
\vspace{-1mm}

\resizebox{0.99\textwidth}{!}
{
\begin{tabular}{lccccc}
\cmidrule[1.0pt]{1-6}
 Task & \multicolumn{2}{c}{Class-Label-to-Image} & \multicolumn{2}{c}{Text-to-Image}& {Super-Resolution}\\
\cmidrule[1.0pt]{1-6}
Dataset $\&$ Resolution & ImageNet 64 & ImageNet 64$\rightarrow$256 & LAION$\&$COYO 64 & LAION$\&$COYO 64$\rightarrow$512 & ImageNet 64$\rightarrow$256\\
\cmidrule[1.0pt]{1-6}
$\vb{z}$ dimension & 64 & 128 & 128& 128 & 128 \\
$\vb{w}$ dimension & 512 & 512 & 1024 & 512 & 512 \\
Adversarial loss type & Logistic & Logistic & Logistic & Logistic & Logistic\\
Conditioning loss type & PD & PD & MS-I/O & MS-I/O & -\\
R1 strength & 0.2048 & 0.2048 & 0.2048 $\sim$ 2.048 & 0.2048 & 0.2048\\
R1 interval & 16 & 16 & 16 & 16 & 16 \\
G Matching loss strength & - & - & 1.0 & 1.0 & - \\
D Matching loss strength & - & - & 1.0 & 1.0 & - \\
LPIPS strength & - & 100.0 & - & 10.0 & 100.0\\
CLIP loss strength & - & - & 0.2 $\sim$ 1.0 & 1.0 & - \\
Optimizer & AdamW  & AdamW & AdamW & AdamW & AdamW\\
Batch size & 1024  & 256 & 512$\sim$1024 & 192$\sim$320 & 256 \\
$G$ learning rate & 0.0025  & 0.0025 & 0.0025 & 0.0025 & 0.0025\\
$D$ learning rate & 0.0025  & 0.0025 & 0.0025 & 0.0025 & 0.0025\\
$\beta_{1}$ for AdamW & 0.0 & 0.0 & 0.0 & 0.0 & 0.0\\
$\beta_{2}$ for AdamW & 0.99 & 0.99 & 0.99 & 0.99 & 0.99\\
Weight decay strength & 0.00001 & 0.00001 & 0.00001 & 0.00001 & 0.00001 \\
Weight decay strength on attention& - & - & 0.01 & 0.01 & - \\
$\#$ $D$ updates per $G$ update & 1 & 1 & 1 & 1 & 1\\
G ema beta & 0.9651 & 0.9912 & 0.9999 & 0.9890 & 0.9912 \\
Precision & TF32 & TF32 & TF32 & TF32 & TF32\\
Mapping Network $M$ layer depth & 2 & 4 & 4 & 4 & 4 \\
Text Transformer $T$ layer depth & - & - & 4 & 2 & - \\
$G$ channel base & 32768 &32768 & 16384 & 32768 & 32768\\
$D$ channel base & 32768 & 32768 & 16384 & 32768 & 32768\\
$G$ channel max & 512 & 512 & 1600 & 512 & 512\\
$D$ channel max & 768 & 512 & 1536 & 512 & 512\\
G $\#$ of filters $N$ for adaptive kernel selection  & 8 & 4 & [1, 1, 2, 4, 8] & [1, 1, 1, 1, 1, 2, 4, 8, 16, 16, 16, 16] & 4 \\
Attention type & self & self & self + cross & self + cross & self \\
$G$ attention resolutions & [8, 16, 32] & [16, 32] & [8, 16, 32] & [8, 16, 32, 64] & [16, 32] \\
$D$ attention resolutions & [8, 16, 32] & - &  [8, 16, 32] & [8, 16] & - \\
$G$ attention depth & [4, 4, 4] & [4, 2] & [2, 2, 1] & [2, 2, 2, 1] & [4, 2] \\
$D$ attention depth & [1, 1, 1] & - & [2, 2, 1] & [2, 2] & - \\
Attention dimension multiplier & 1.0 & 1.4 & 1.0 & 1.0 & 1.4 \\
MLP dimension multiplier of attention & 4.0 & 4.0 & 4.0 & 4.0 & 4.0 \\
$\#$ synthesis block per resolution & 1 & 5 & [3, 3, 3, 2, 2] & [4, 4, 4, 4, 4, 4, 3] & 5 \\
$\#$ discriminator block per resolution & 1 & 1 & [1, 2, 2, 2, 2] & 1 & - \\
Residual gain & 1.0 & 0.4 & 0.4 & 0.4 & 0.4 \\
Residual gain on attention & 1.0 & 0.3 & 0.3 & 0.5 & 0.3 \\
MinibatchStdLayer & True & True & False & True & True\\
D epilogue mbstd group size & 8 & 4 & - & 2 & 4 \\
Multi-scale training & False & False & True & True & False \\
Multi-scale loss ratio (high to low res) & - & - & [0.33, 0.17, 0.17, 0.17, 0.17] &  & - \\
D intermediate layer adv loss weight& - & - & 0.01 & [0.2, 0.2, 0.1, 0.1, 0.1, 0.1, 0.1, 0.1] & - \\
D intermediate layer matching loss weight & - & - & 0.05 & - & - \\
Vision-aided discriminator backbone & - & - & CLIP-ViT-B/32-V & - & -\\
\cmidrule[1.0pt]{1-6}
$G$ Model size & 209.5M & 359.8M & 652.5M & 359.1M & 359.0M\\
$D$ Model size & 76.7M & 30.7M & 381.4M & 130.1M & 28.9M\\
Iterations & 300k & 620k & 1350k & 915k & 160k \\
$\#$ A100 GPUs for training & 64  & 64 & 96$\sim$128 & 64 & 32 \\
\cmidrule[1.0pt]{1-6}
\end{tabular}}
\label{table:details}
\vspace{-2mm}
\end{table*}
\end{center}

\pagebreak

\clearpage

\section{Text-to-image synthesis results}
\lblsec{C}
\vspace{8pt}
\subsection{Truncation trick at inference}

Similar to the classifier guidance~\cite{dhariwal2021diffusion} and classifier-free guidance~\cite{ho2022classifier} used in diffusion models such as LDM, our GAN model can leverage the truncation trick~\cite{Brock2019LargeSG,karras2019style} at inference time. 
\begin{equation}
\vb{w_{\text{trunc}}} = \text{lerp}(\vb{w_\text{mean}}, \vb{w}, \psi),
\end{equation}

\noindent where $\vb{w_\text{mean}}$ is the mean of $\vb{w}$ of the entire dataset, which can be precomputed. In essence, the truncation trick lets us trade diversity for fidelity by interpolating the latent vector to the mean of the distribution and thereby making the outputs more typical. When $\psi = 1.0$, $\vb{w_\text{mean}}$ is not used, and there is no truncation. When $\psi = 0.0$, $\vb{w}$ collapses to the mean, losing diversity.

\begin{figure*}[ht]
    \centering
    \includegraphics[width=1.0\linewidth]{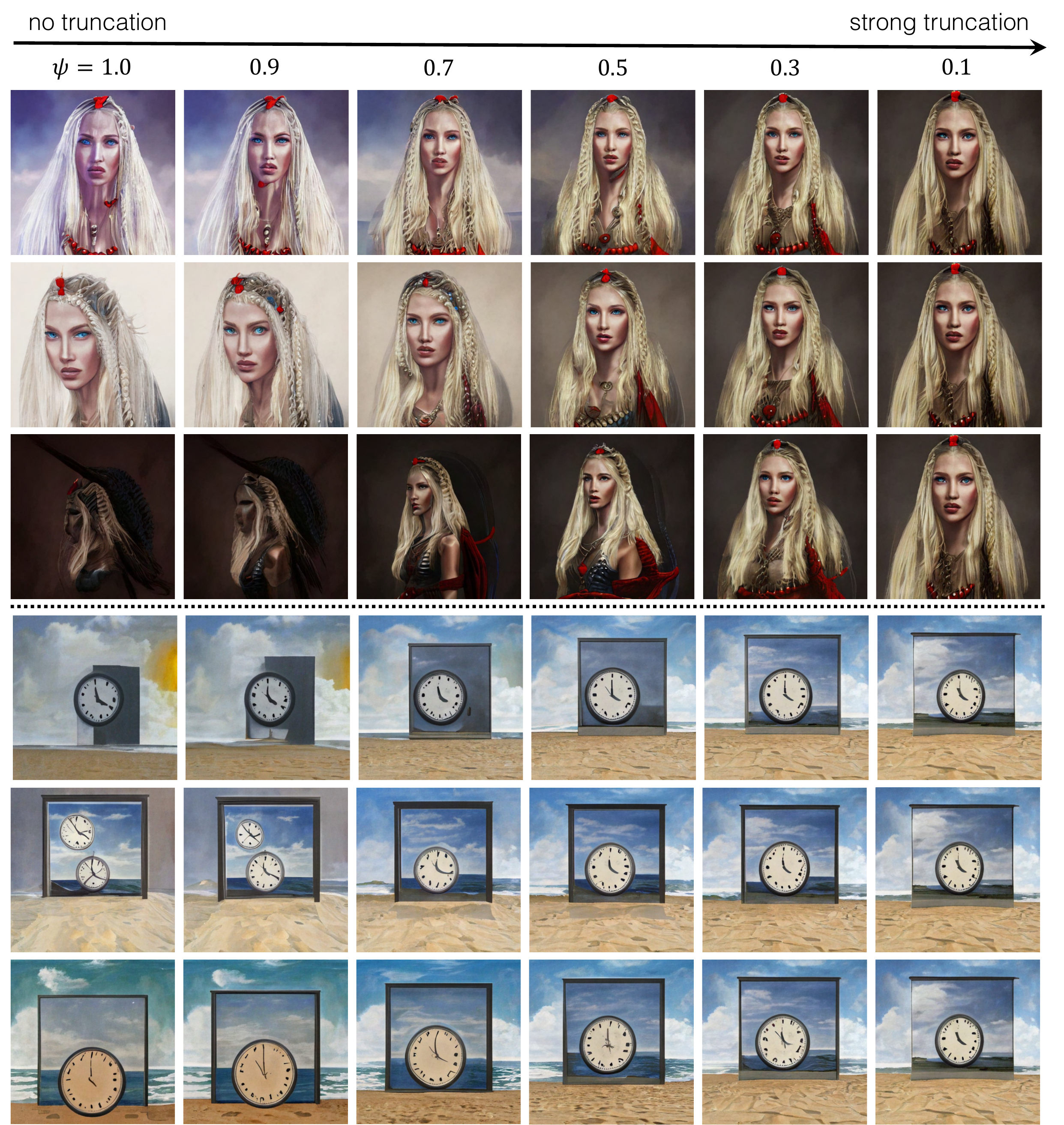}
    \caption{{\bf The visual effect of our truncation trick}. We demonstrate the effect of truncation by decreasing the truncation value $\psi$ from 1.0. We show six example outputs with the text prompt ``digital painting of a confident and severe looking northern war goddess, extremely long blond braided hair, beautiful blue eyes and red lips." and ``Magritte painting of a clock on a beach.". At 1.0 (no truncation), the diversity is high, but the alignment is not satisfactory. As the truncation increases, text-image alignment improves, at the cost of diversity. We find that a truncation value between 0.8 and 0.7 produces the best result. 
    }
    \label{fig:suppmat_truncation}
\end{figure*}

While it is straightforward to apply the truncation trick for the unconditional case, it is less clear how to achieve this for text-conditional image generation. We find that interpolating the latent vector toward both the mean of the entire distribution as well as the mean of $\vb{w}$ conditioned on the text prompt produces desirable results. 

\begin{equation}
\vb{w_{\text{trunc}}} = \text{lerp}(\vb{w_{\text{mean}, \vb{c}}}, \text{lerp}( \vb{w_\text{mean}}, \vb{w}, \psi ), \psi ),
\end{equation}

\noindent where $\vb{w}_{\text{mean}, \vb{c}}$ can be computed at inference time by sampling $\vb{w} = M(\vb{z}, \vb{c})$ 16 times with the same $\vb{c}$, and taking the average. This operation's overhead is negligible, as the mapping network $M$ is computationally light compared to the synthesis network. At $\psi = 1.0$, $\vb{w_{\text{trunc}}}$ becomes $\vb{w_{\text{trunc}}} = \vb{w}$, meaning no truncation.  Figure~\ref{fig:suppmat_truncation} demonstrates the effect of our text-conditioned truncation trick. 

\begin{figure}[b]
    \centering
    \includegraphics[width=0.94\linewidth]{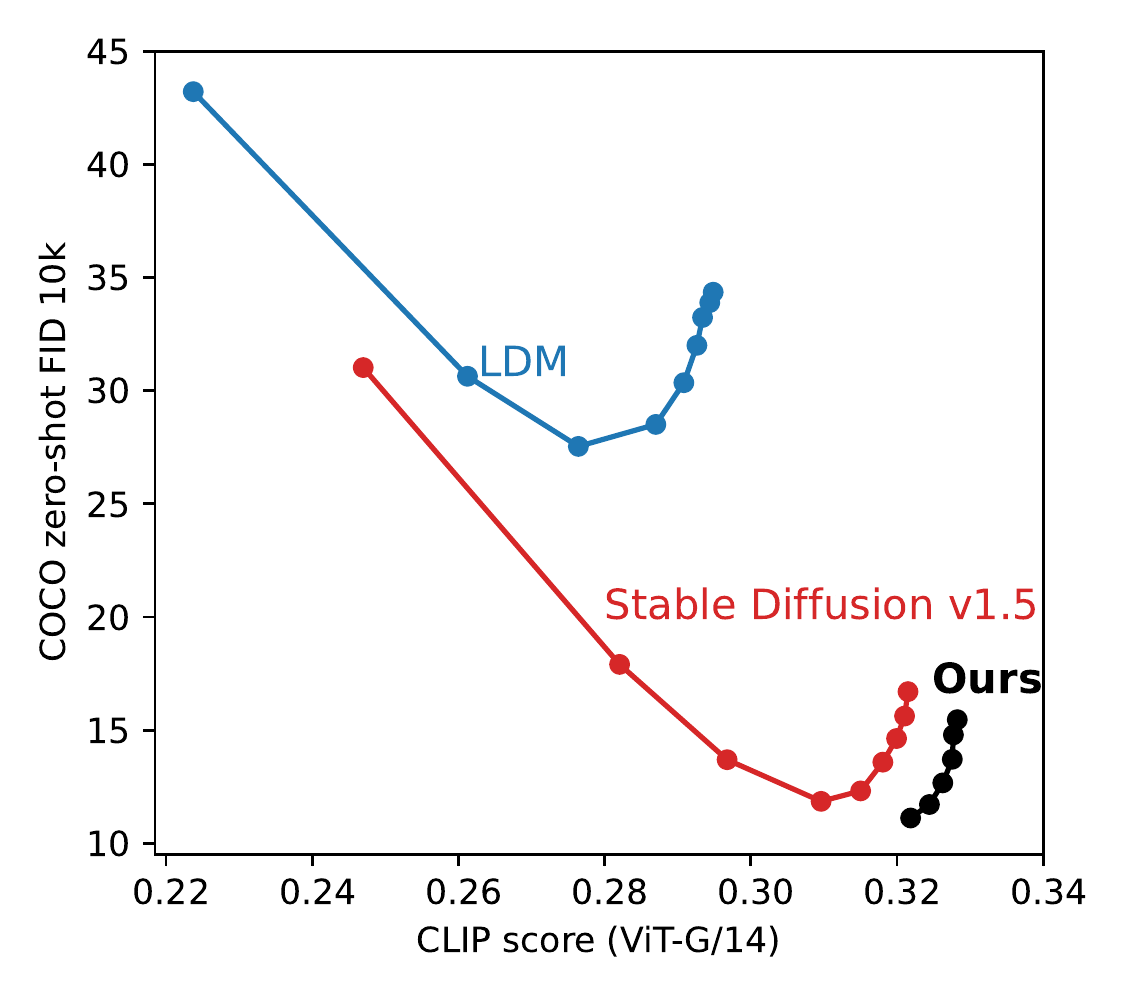}
    \vspace{-4mm}
    \caption{ We investigate how our FID and CLIP score changes over different truncation values [1.0, 0.9, 0.8, 0.7, 0.6, 0.5], by visualizing them along with the FID-CLIP score curve of two publicly available large scale diffusion models: LDM and Stable Diffusion. It is seen that the CLIP score increases with more truncation, at the cost of reduced diversity indicated by higher FID. The guidance values of the diffusion models are [1.0, 1.25, 1.5, 1.75, 2, 4, 6, 8, 10].}
    \label{fig:suppmat_truncation_plot}
\end{figure}

Quantitatively, the effect of truncation is similar to the guidance technique of diffusion models. As shown in  Figure~\ref{fig:suppmat_truncation_plot}, the CLIP score increases with more truncation, where the FID increases due to reduced diversity. 

\subsection{Comparison to diffusion models}

Finally, we show randomly sampled results of our model and compare them with publicly available diffusion models, LDM~\cite{rombach2022high}, Stable Diffusion~\cite{stablediffusion}, and DALL$\cdot$E 2~\cite{ramesh2022hierarchical}.

\begin{figure*}[t!]
    \centering
    \includegraphics[width=0.99\linewidth]{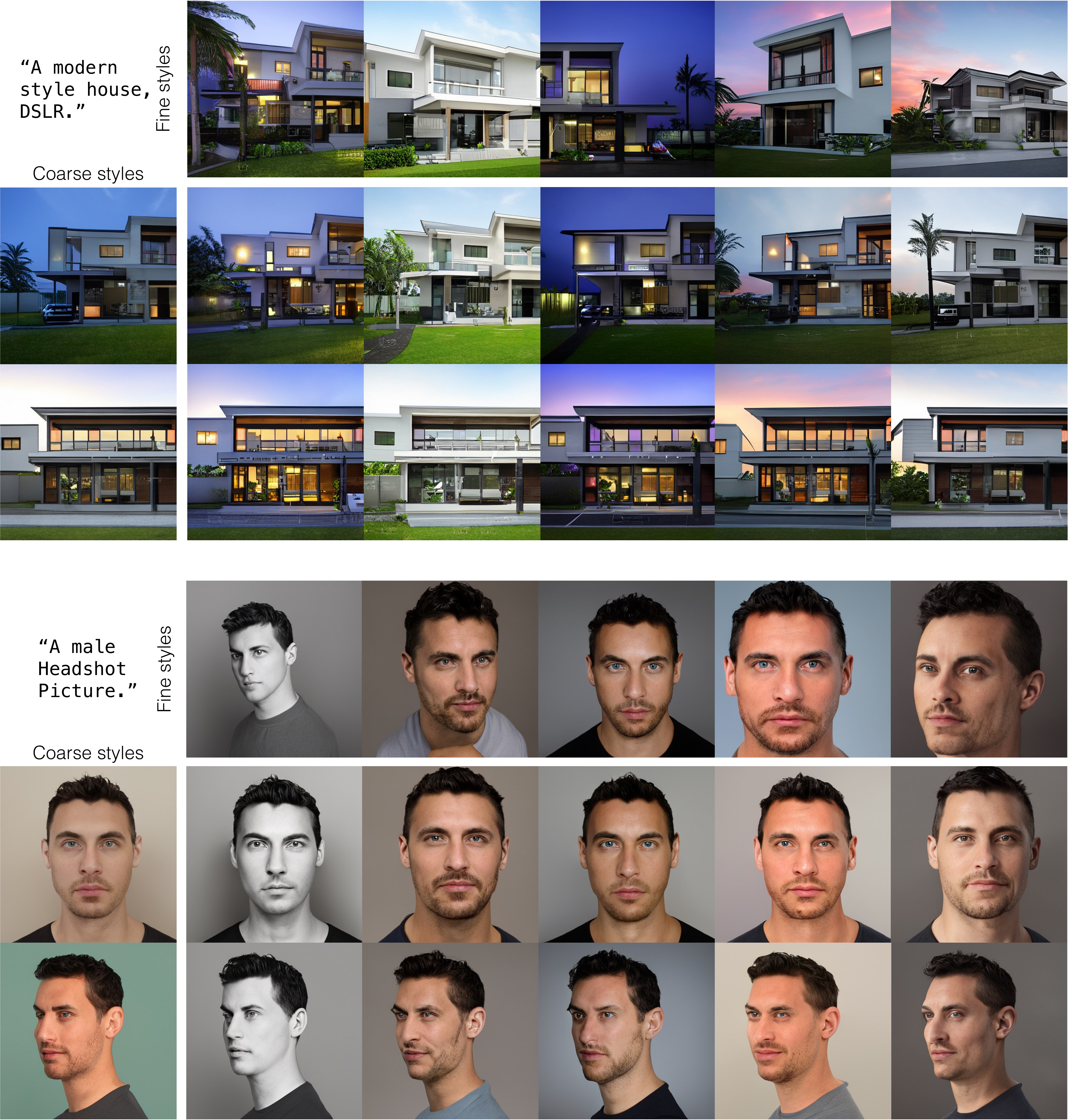}
    \caption{
    {\bf Style mixing.} GigaGAN maintains a disentangled latent space, allowing us to blend the coarse style of one sample with the fine style of another. The corresponding latent codes are spliced together to produce a style-swapping grid. The outputs are generated from the same prompt but with different latent codes.}
    \label{fig:suple_stylization2}
\end{figure*}
\begin{figure*}[t!]
    \centering
    \includegraphics[width=0.99\linewidth]{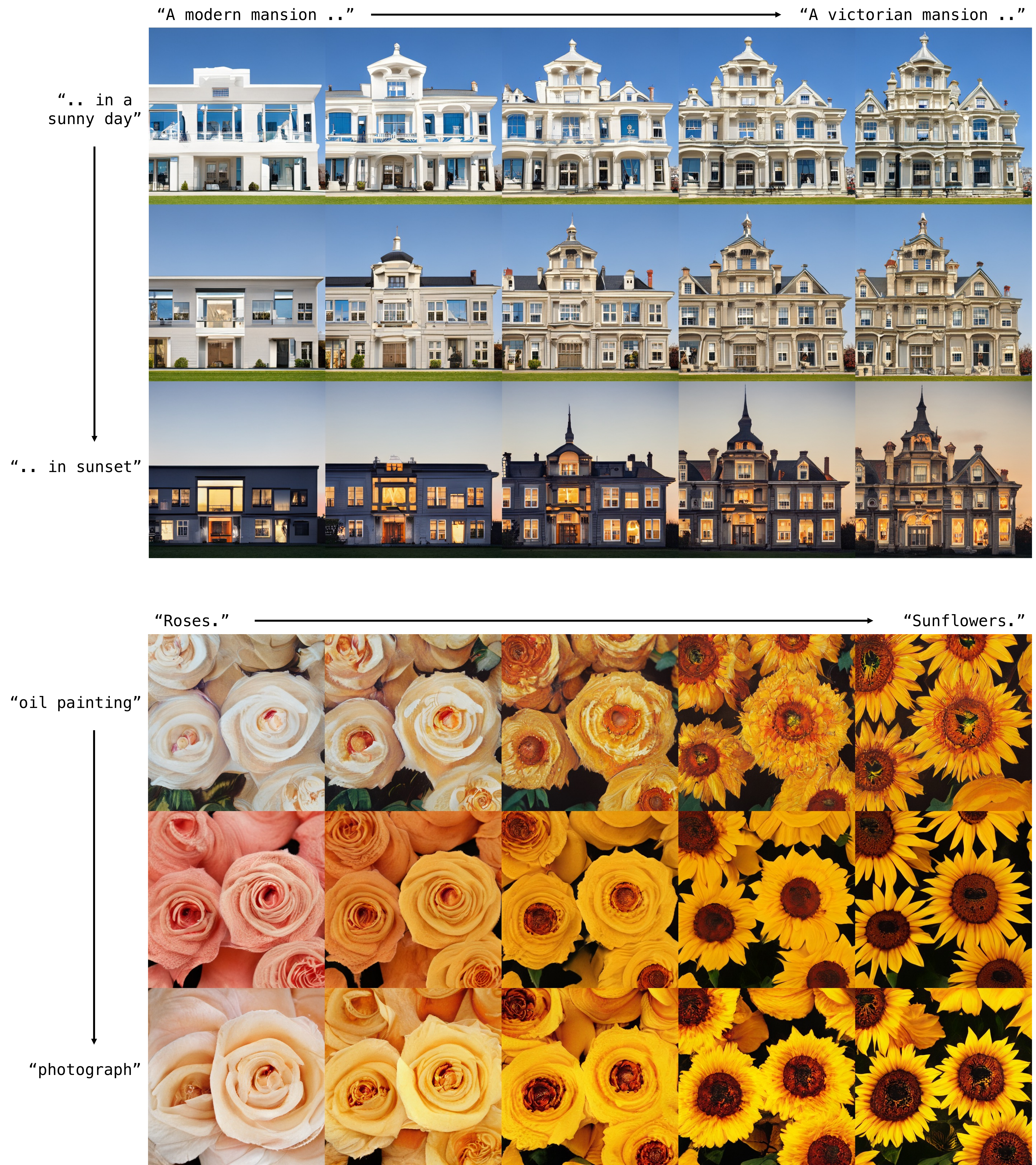}
    \captionsetup{width=.97\linewidth}
    \captionsetup{singlelinecheck = false, justification=justified}
    \caption{
    {\bf Prompt interpolation}. GigaGAN enables smooth interpolation between prompts, as shown in the interpolation grid. The four corners are generated from the same latent but with different text prompts. The corresponding text embeddings  and style vectors are interpolated to create a smooth transition. The same  results in similar layouts.
  }
    \label{fig:interpolation}
\end{figure*}
   
\begin{figure*}[t!]
    \centering
    \includegraphics[width=0.99\linewidth]{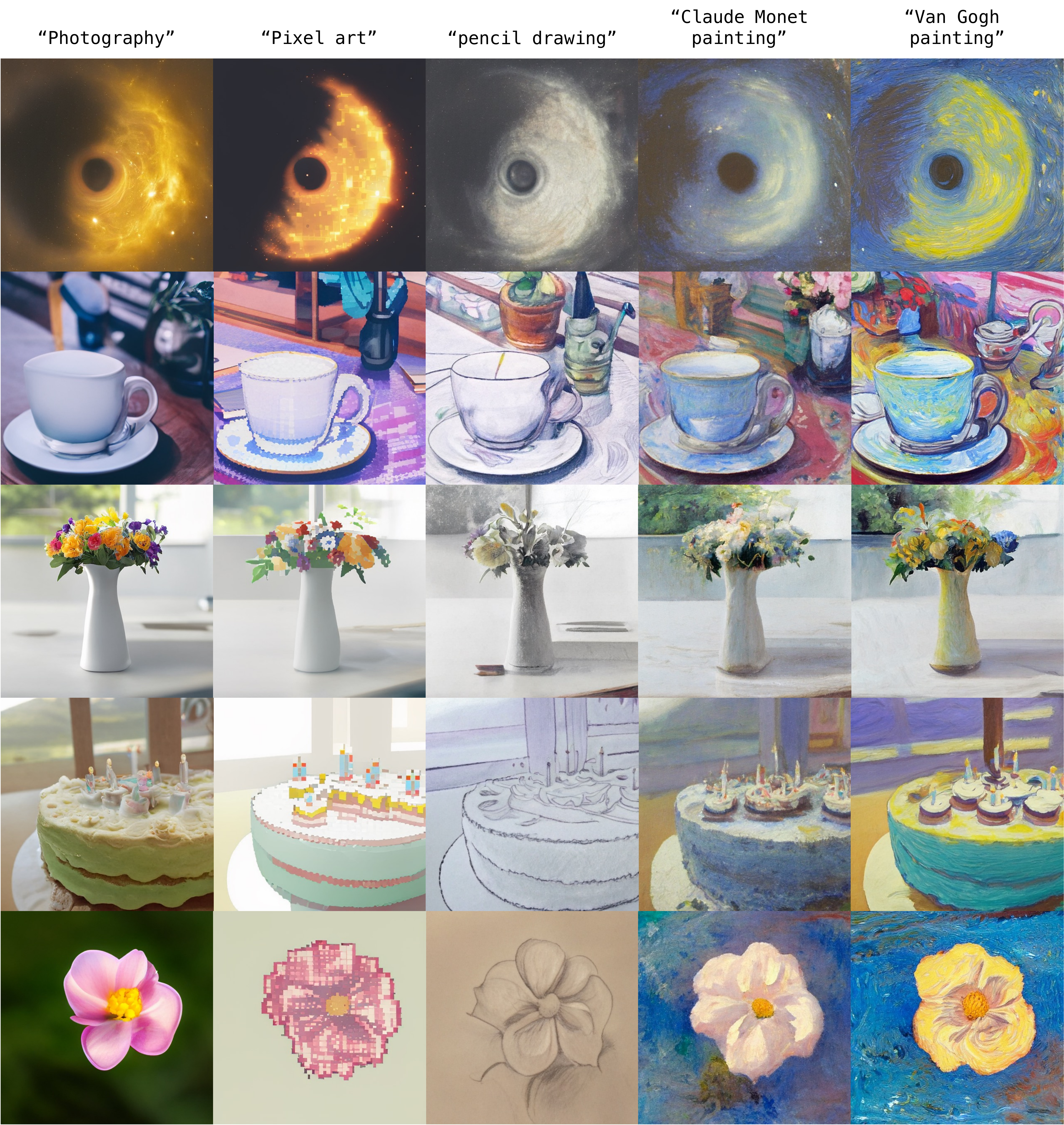}
    \captionsetup{width=.97\linewidth}
    \captionsetup{singlelinecheck = false, justification=justified}
    \caption{
    {\bf Prompt mixing}. GigaGAN can directly control the style with text prompts. Here we generate five outputs using the prompts ``Photography of X", shown in the ``Photography'' column. Then we re-compute the text embeddings $\vb{t}$ and the style codes $\vb{w}$ using the new prompts ``Y of X", such as ``Van Gogh painting of the black hole in the space", and apply them to the second half layers of the generator, achieving layout-preserving style control. Cross-attention mechanism automatically localizes the style to the object of interest. We use
the following prompts in order from the row above. (1) the black hole in the space. (2) a teacup on the desk. (3) a table top with a vase of flowers on it. (4) a birthday cake. (5) a beautiful flower. We discover that GigaGAN's prompt-based style transfer is only possible for images of a single and simple object.}
    \label{fig:style_transfer}
\end{figure*}
\begin{figure*}[t!]
    \centering
    \includegraphics[width=0.99\linewidth]{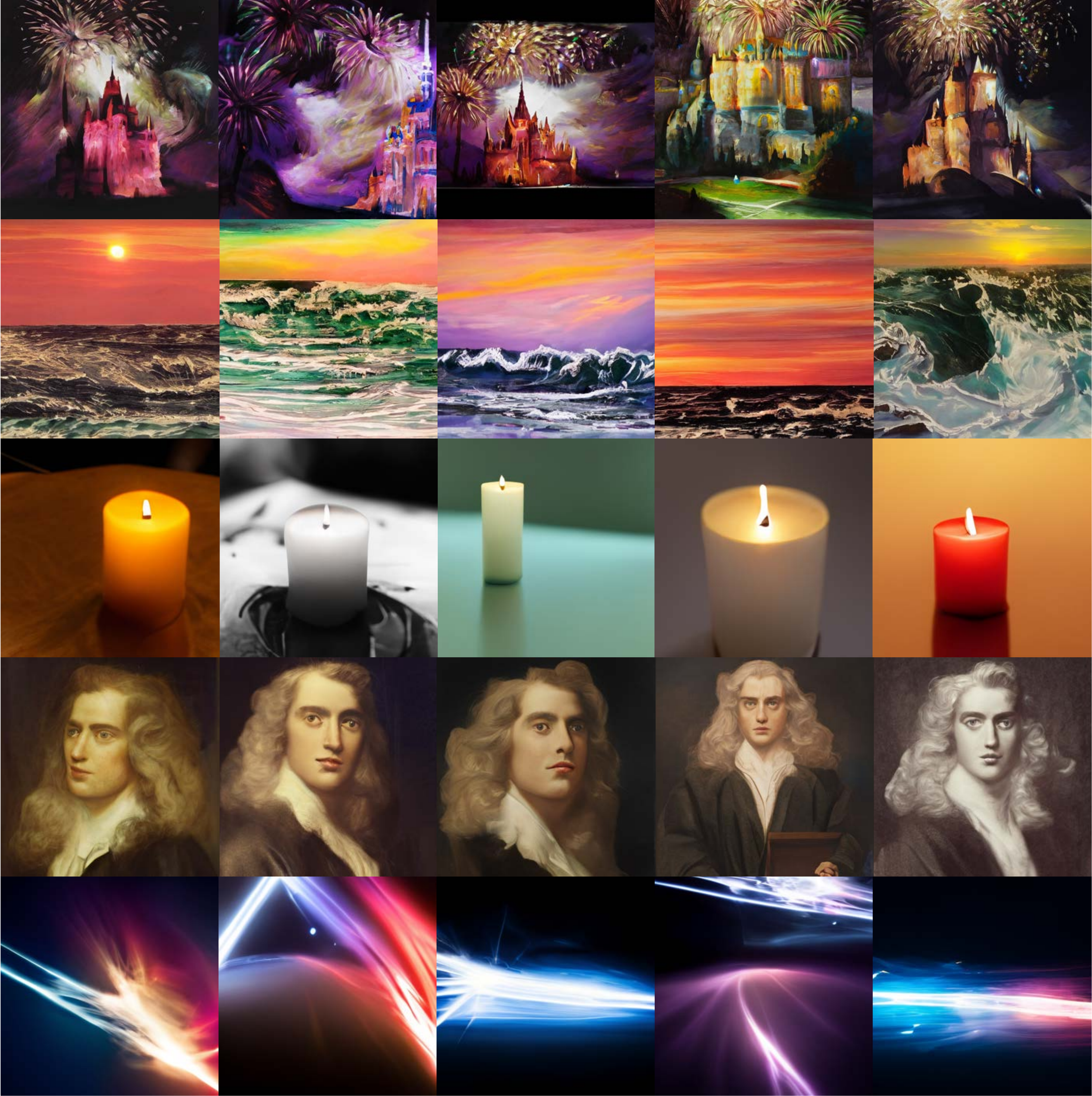}
    \captionsetup{width=.97\linewidth}
    \captionsetup{singlelinecheck = false, justification=justified}
    \caption{
    {\bf Diversity of synthesized images using GigaGAN}. GigaGAN can synthesize diverse images for a given prompt. We use the following prompts in order from the row above. (1) Majestic castle and fireworks, art work, oil painting, 2k. (2) Oil-painting depicting a sunset over the sea with waves. (3) A burning candle with tho wicks, detailed photo, studio lighting. (4) Portrait of Isaac Newton, long hair. (5) An abstract representation of the speed of light.
  }
    \label{fig:different_zs}
\end{figure*}

\begin{figure*}[t!]
    \centering
    \includegraphics[width=1.0\linewidth]{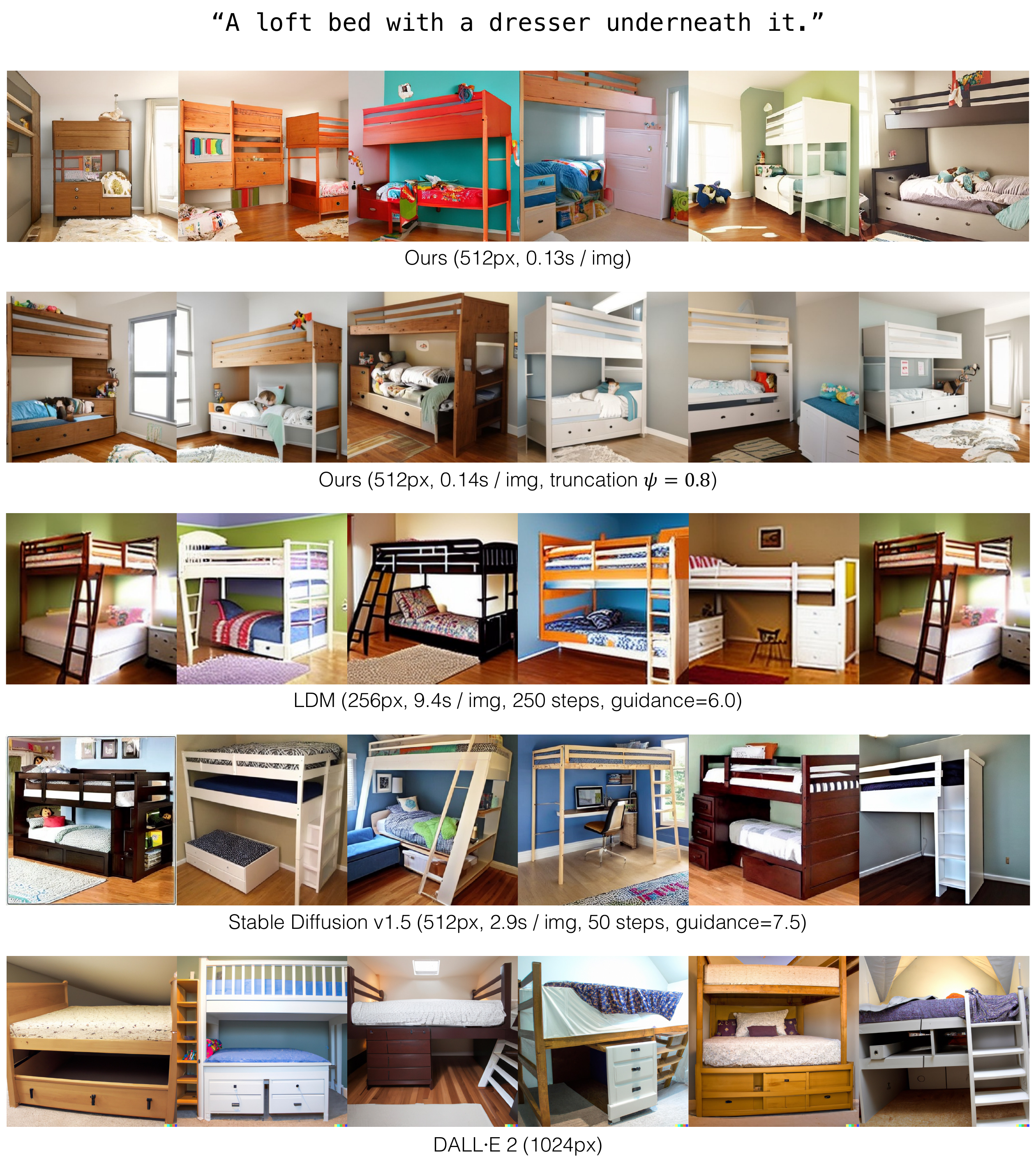}
    \vspace{-7mm}
    \caption{Random outputs of our model, Latent Diffusion Model~\cite{rombach2022high}, Stable Diffusion~\cite{stablediffusion}, and DALL$\cdot$E 2~\cite{ramesh2022hierarchical}, using prompt ``A loft bed with a dresser underneath it". We show two versions of our model, one without truncation and the other with truncation. Our model enjoys faster speed than the diffusion models. Still, we observe our model falls behind in structural coherency, such as the number of legs of the bed frames. For LDM and Stable Diffusion, we use 250 and 50 sampling steps with DDIM / PLMS~\cite{liu2022pseudo}, respectively. For DALL$\cdot$E 2, we generate images using the official DALL$\cdot$E service~\cite{dalle2api}.
    }
    \label{fig:suppmat_bed}
\end{figure*}

\begin{figure*}[t!]
    \centering
    \includegraphics[width=1.0\linewidth]{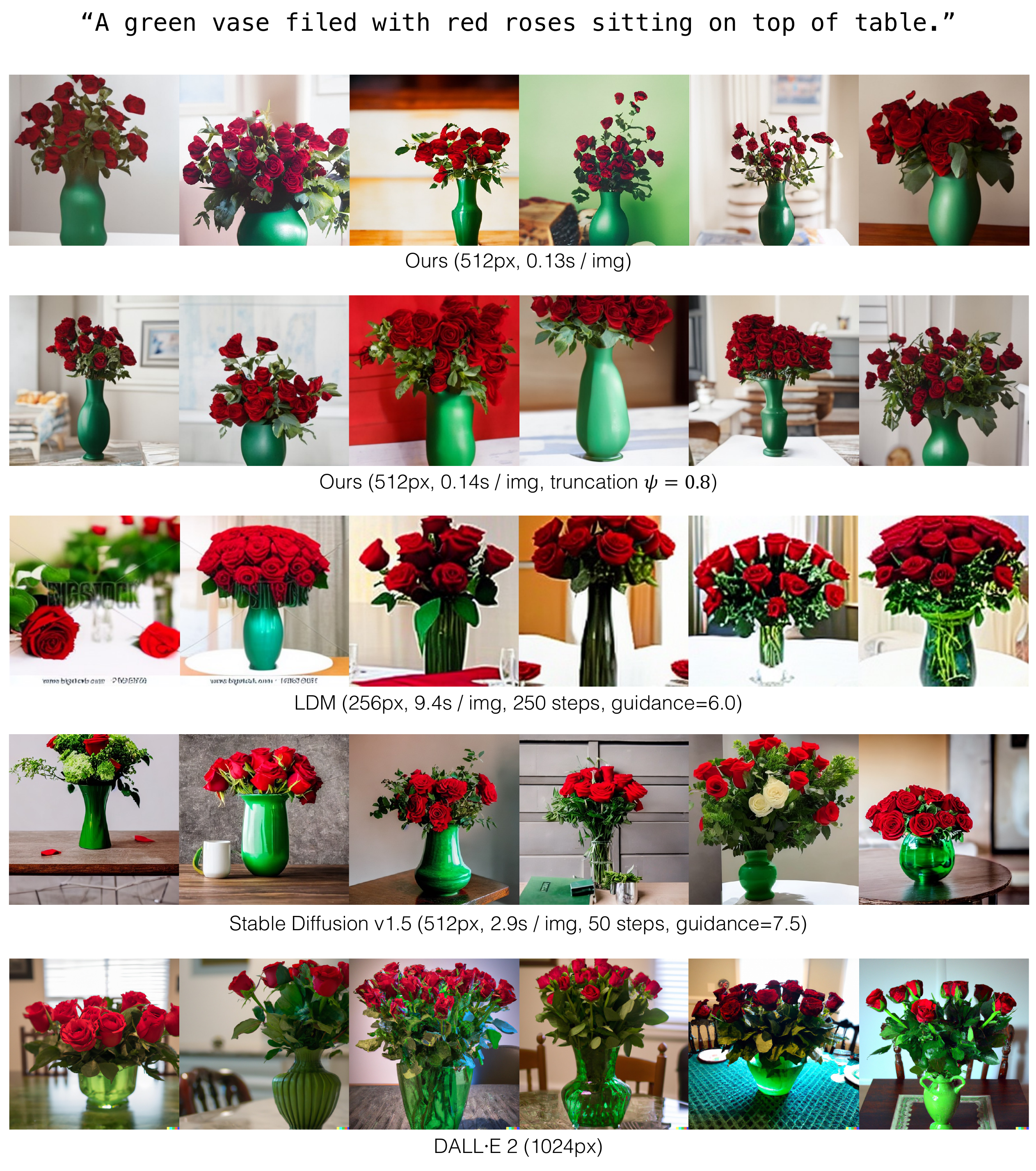}
    \caption{Random outputs of our model, Latent Diffusion Model~\cite{rombach2022high}, Stable Diffusion~\cite{stablediffusion}, and DALL$\cdot$E 2~\cite{ramesh2022hierarchical}, using prompt ``A green vase filed with red roses sitting on top of table". We show two versions of our model, one without truncation and the other with truncation. Our model enjoys faster speed than the diffusion models in both cases. Still, we observe our model falls behind in structural coherency like the symmetry of the vases. For LDM and Stable Diffusion, we use 250 and 50 sampling steps with DDIM / PLMS~\cite{liu2022pseudo}, respectively. For DALL$\cdot$E 2, we generate images using the official DALL$\cdot$E service~\cite{dalle2api}.
    }
    \label{fig:suppmat_flower}
\end{figure*}

\begin{figure*}[t!]
    \centering
    \includegraphics[width=1.0\linewidth]{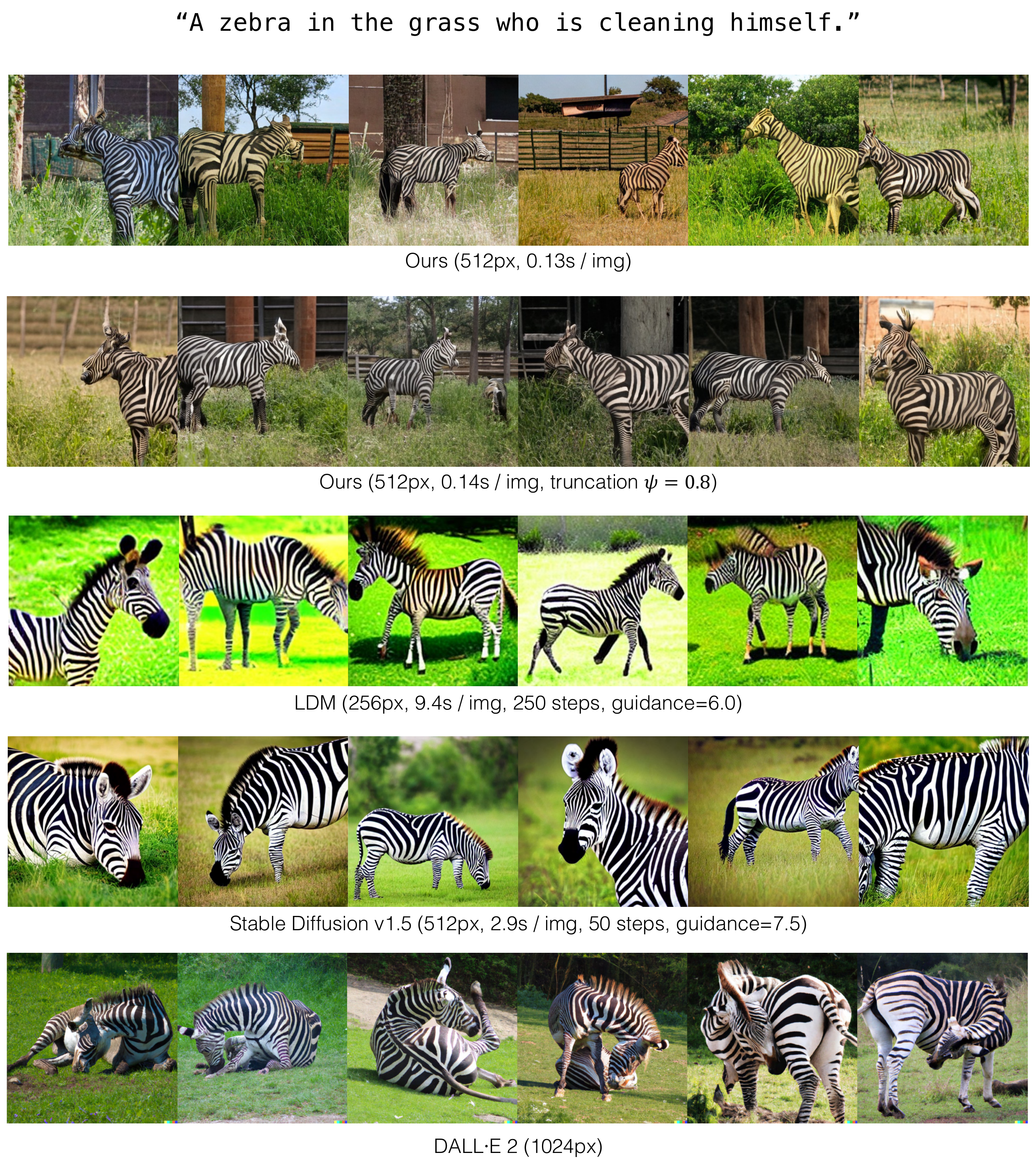}
    \caption{Random outputs of our model, Latent Diffusion Model~\cite{rombach2022high}, Stable Diffusion~\cite{stablediffusion}, and DALL$\cdot$E 2~\cite{ramesh2022hierarchical}, using prompt ``A zebra in the grass who is cleaning himself". We show two versions of our model, one without truncation and the other with truncation. Our model enjoys faster speed than the diffusion models in both cases. Still, we observe our model falls behind in details, such as the precise stripe pattern of the positioning of eyes. For LDM and Stable Diffusion, we use 250 and 50 sampling steps with DDIM / PLMS~\cite{liu2022pseudo}, respectively. For DALL$\cdot$E 2, we generate images using the official DALL$\cdot$E service~\cite{dalle2api}.
    }
    \label{fig:suppmat_zebra}
\end{figure*}

\begin{figure*}[t!]
    \centering
    \includegraphics[width=1.0\linewidth]{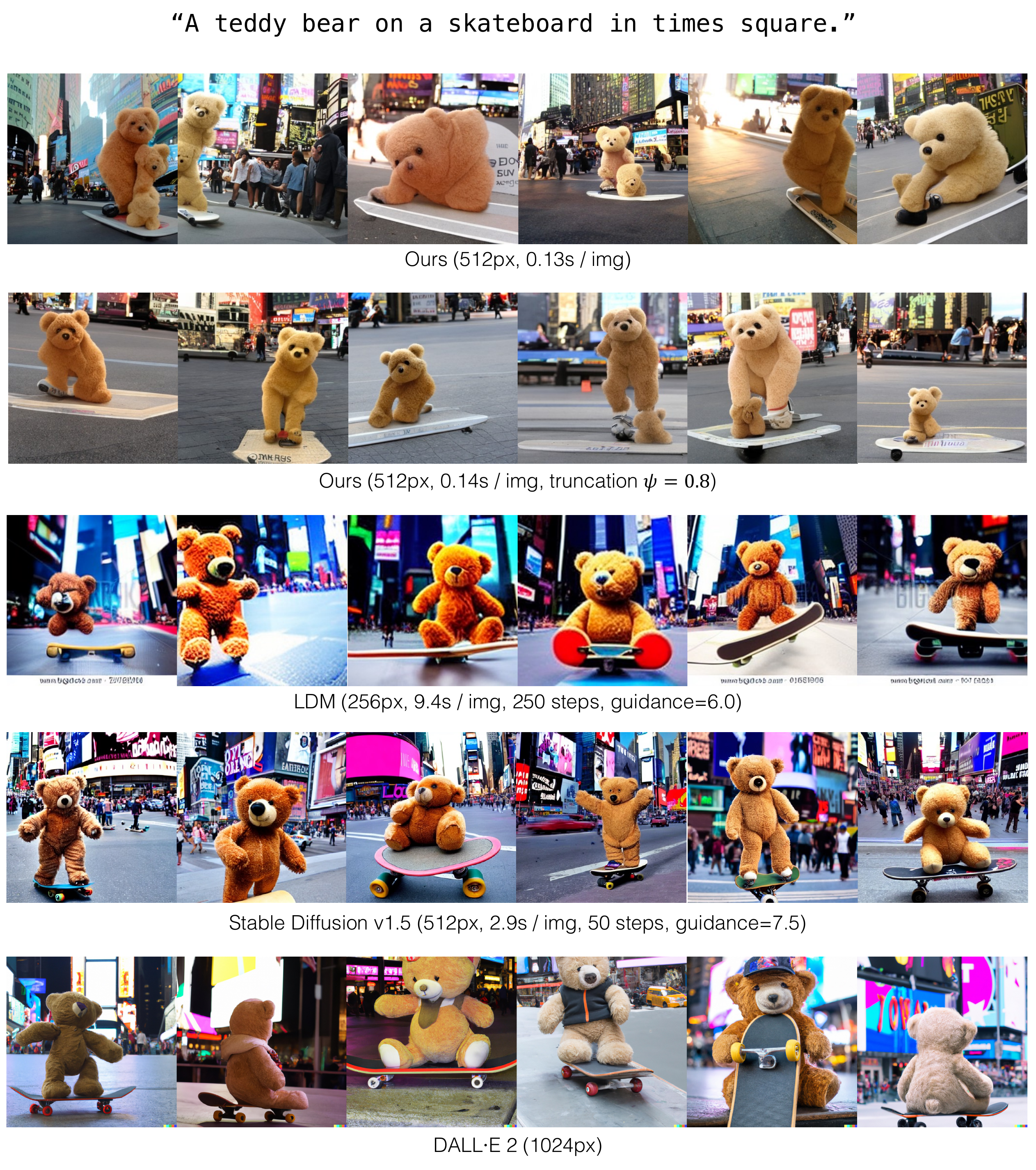}
    \caption{Random outputs of our model, Latent Diffusion Model~\cite{rombach2022high}, Stable Diffusion~\cite{stablediffusion}, and DALL$\cdot$E 2~\cite{ramesh2022hierarchical}, using prompt ``A teddy bear on a skateboard in times square". We show two versions of our model, one without truncation and the other with truncation. Our model enjoys faster speed than the diffusion models in both cases. Still, we observe our model falls behind in details, like the exact shape of skateboards. For LDM and Stable Diffusion, we use 250 and 50 sampling steps with DDIM / PLMS~\cite{liu2022pseudo}, respectively. For DALL$\cdot$E 2, we generate images using the official DALL$\cdot$E service~\cite{dalle2api}.
    }
    \label{fig:suppmat_timesquare}
\end{figure*}

\begin{figure*}[t!]
    \centering
    \includegraphics[width=1.0\linewidth]{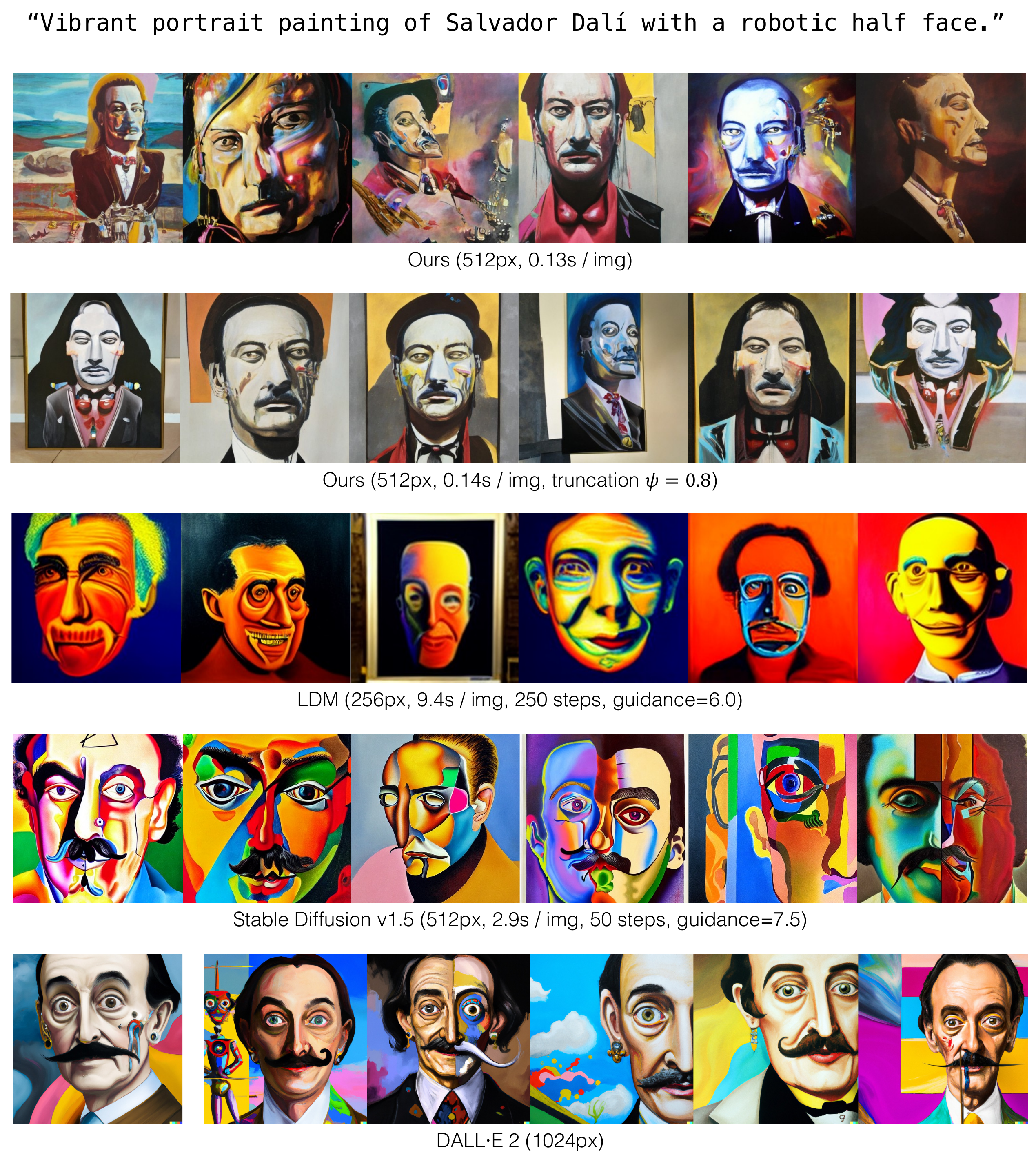}
    \caption{Random outputs of our model, Latent Diffusion Model~\cite{rombach2022high}, Stable Diffusion~\cite{stablediffusion}, and DALL$\cdot$E 2~\cite{ramesh2022hierarchical}, using prompt ``Vibrant portrait painting of Salvador Dalí with a robotic half face". We show two versions of our model, one without truncation and the other with truncation. Our model enjoys faster speed than the diffusion models in both cases. Still, we observe our model falls behind in structural details like in the detailed shape of eyes. For LDM and Stable Diffusion, we use 250 and 50 sampling steps with DDIM / PLMS~\cite{liu2022pseudo}, respectively. For DALL$\cdot$E 2, we generate images using the official DALL$\cdot$E service~\cite{dalle2api}.
    }
    \label{fig:suppmat_hamburger_dalle}
\end{figure*}

\begin{figure*}[t!]
    \centering
    \includegraphics[width=1.0\linewidth]{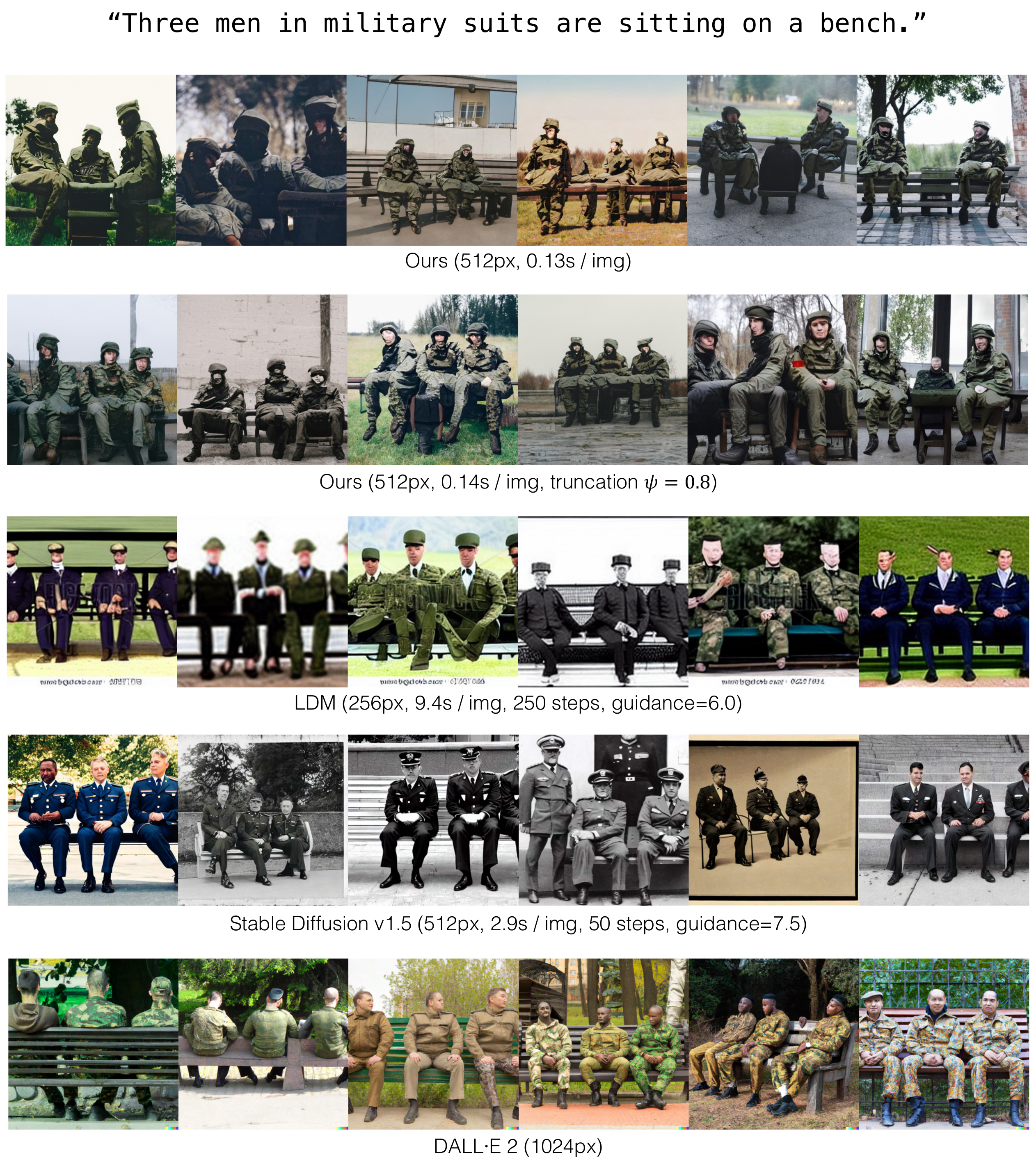}
    \caption{Random outputs of our model, Latent Diffusion Model~\cite{rombach2022high}, Stable Diffusion~\cite{stablediffusion}, and DALL$\cdot$E 2~\cite{ramesh2022hierarchical}, using prompt ``Three men in military suits are sitting on a bench". We show two versions of our model, one without truncation and the other with truncation. Our model enjoys faster speed than the diffusion models in both cases. Still, we observe our model falls behind in details in facial expression and attire. For LDM and Stable Diffusion, we use 250 and 50 sampling steps with DDIM / PLMS~\cite{liu2022pseudo}, respectively. For DALL$\cdot$E 2, we generate images using the official DALL$\cdot$E service~\cite{dalle2api}.
    }
    \label{fig:suppmat_hamburger_military}
\end{figure*}

\begin{figure*}[!ht]
    \centering
    
    \includegraphics[width=0.97\linewidth]{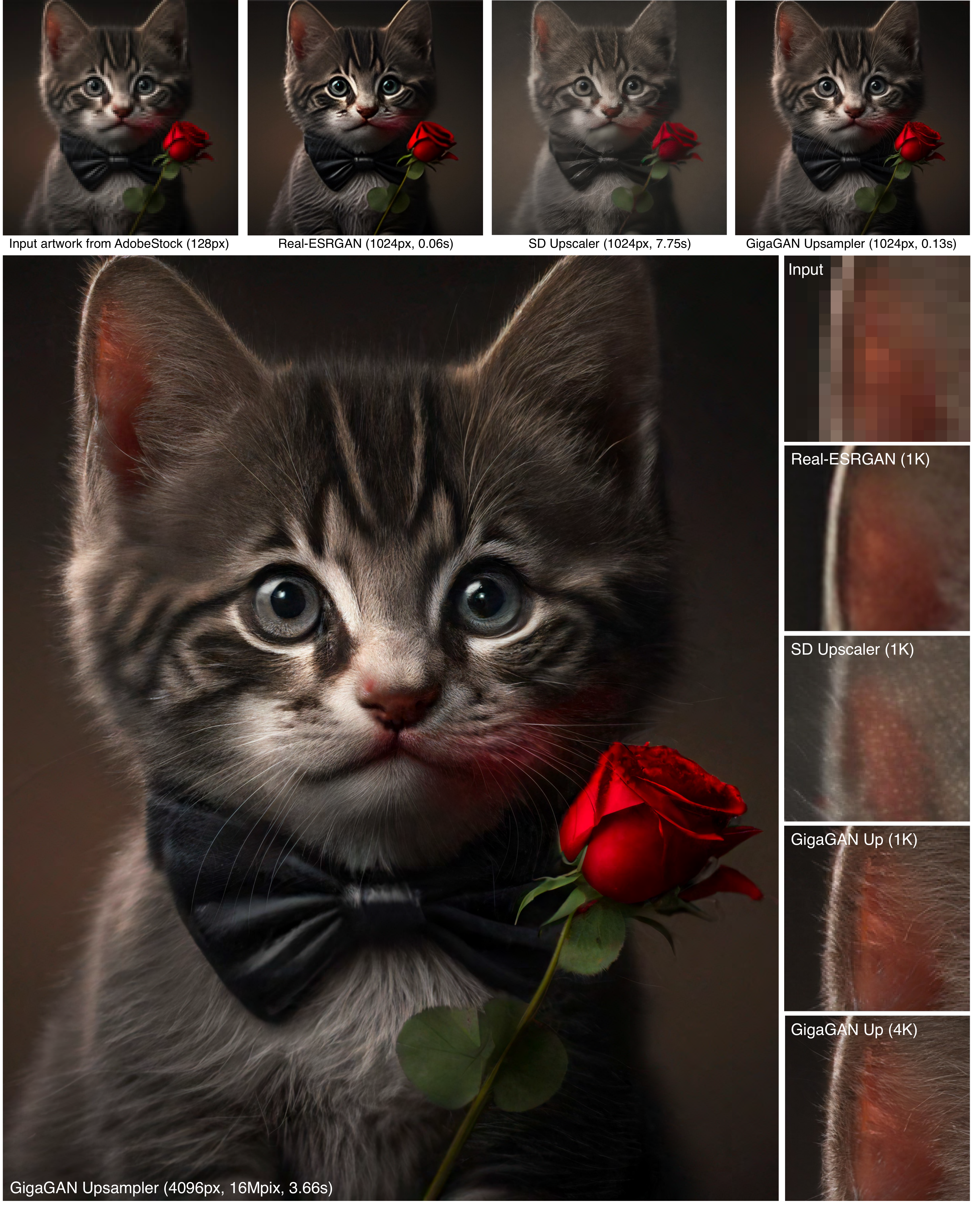}
    \vspace{-3mm}
    \caption{{\bf Our GAN-based upsampler} can serve as the upsampler for many text-to-image models that generate initial outputs at low resolutions like 64px or 128px. We simulate such usage by applying our 8$\times$ superresolution model on a low-res 128px artwork to obtain the 1K output, using ``Portrait of a kitten dressed in a bow tie. Red Rose. Valentine's day.". Then our model can be re-applied to go beyond 4K. We compare our model with the text-conditioned upscaler of Stable Diffusion~\cite{stablediffusion} and unconditional Real-ESRGAN~\cite{wang2021realesrgan}. Zooming in is recommended for comparison between 1K and 4K outputs. }
    \vspace{-2mm}
    \label{fig:superres_cat}
\end{figure*}

\begin{figure*}[!ht]
    \centering

    \includegraphics[width=0.97\linewidth]{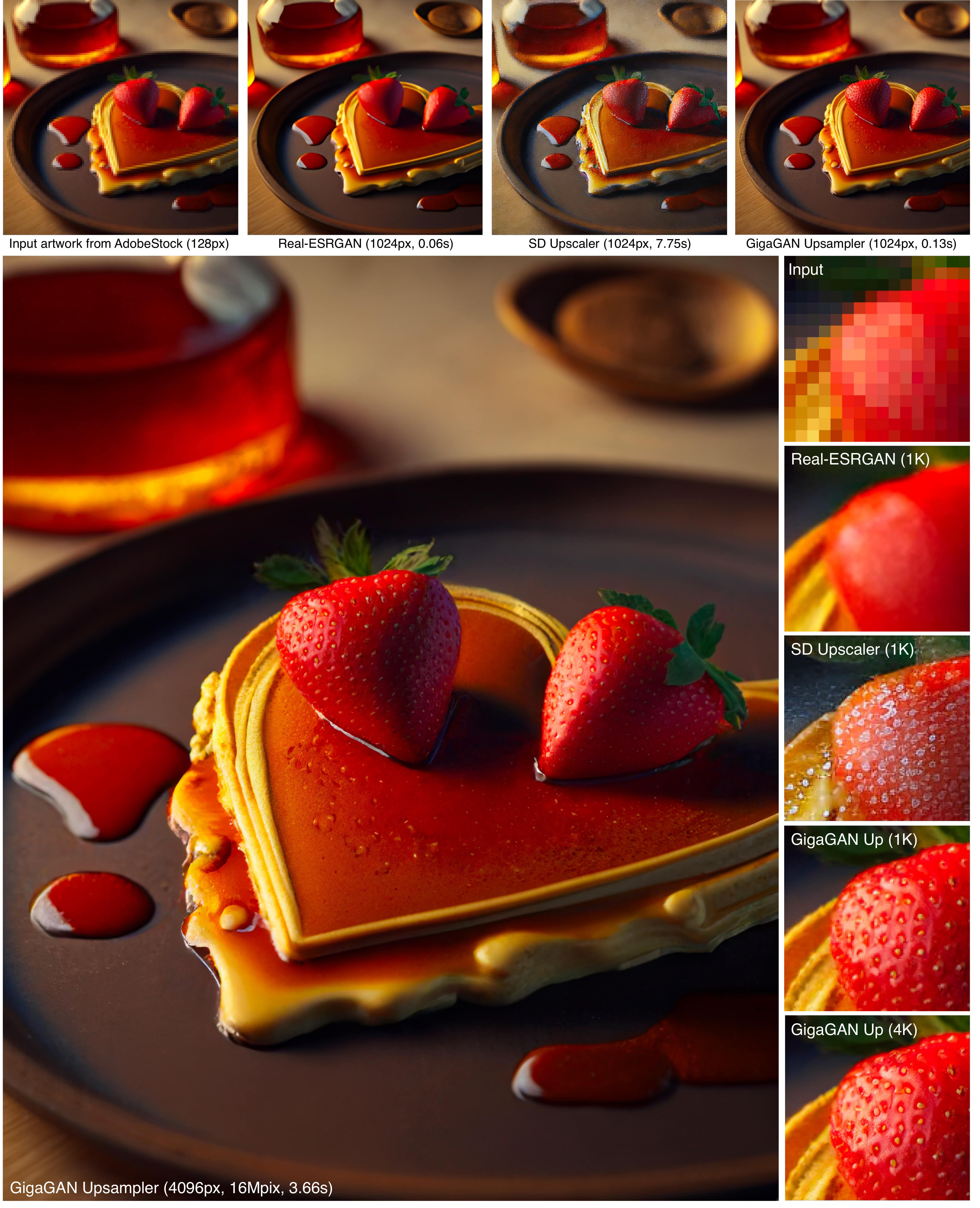}
    \vspace{-3mm}
    \caption{{\bf Our GAN-based upsampler} can serve as the upsampler for many text-to-image models that generate initial outputs at low resolutions like 64px or 128px. We simulate such usage by applying our 8$\times$ superresolution model on a low-res 128px artwork to obtain the 1K output, using ``Heart shaped pancakes with honey and strawberry for Valentine's Day". Then our model can be re-applied to go beyond 4K. We compare our model with the text-conditioned upscaler of Stable Diffusion~\cite{stablediffusion} and unconditional Real-ESRGAN~\cite{wang2021realesrgan}. Zooming in is recommended for comparison between 1K and 4K outputs.}
    \vspace{-2mm}
    \label{fig:superres_pancake}
\end{figure*}

\begin{figure*}[!ht]
    \centering
    \includegraphics[width=0.97\linewidth]{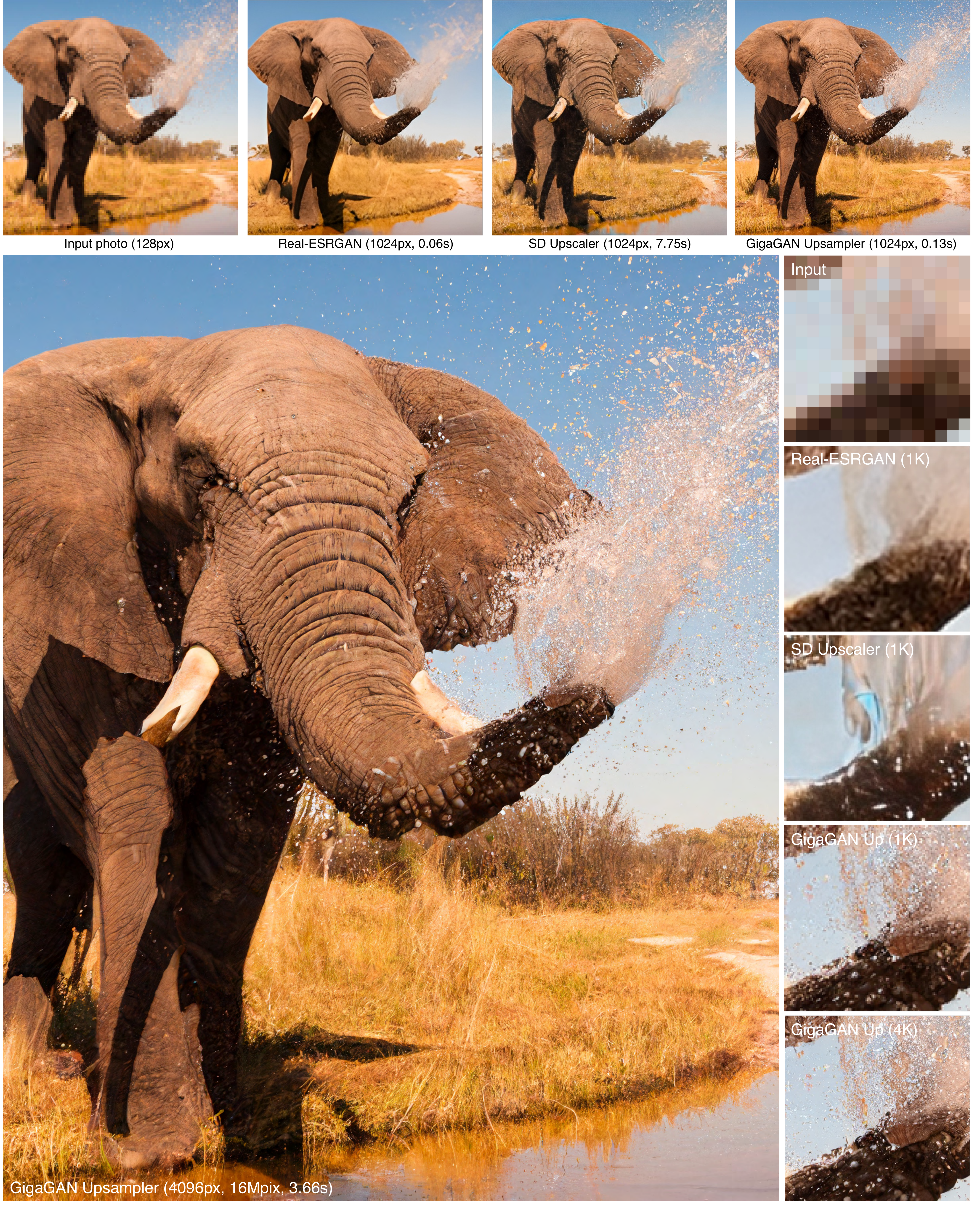}
    \vspace{-3mm}
    \caption{{\bf Our GAN-based upsampler} can also be used as an off-the-shelf superresolution model for real images with a large scaling factor by providing an appropriate description of the image. We apply our text-conditioned 8$\times$ superresolution model on a low-res 128px photo to obtain the 1K output, using ``An elephant spraying water with its trunk". Then our model can be re-applied to go beyond 4K. We compare our model with the text-conditioned upscaler of Stable Diffusion~\cite{stablediffusion} and unconditional Real-ESRGAN~\cite{wang2021realesrgan}. Zooming in is recommended for comparison between 1K and 4K outputs.}
    \vspace{-2mm}
    \label{fig:superres_elephant}
\end{figure*}

\clearpage
\end{document}